%% file: main.tex
\documentclass[11pt]{article}

\usepackage{amsmath}
\usepackage{amsthm}
\usepackage{bm}

\usepackage[final]{automl}

\input{math_commands.tex}

\usepackage{microtype} 
\usepackage{booktabs}  

\usepackage{hyperref}       
\usepackage{url}            
\hypersetup{colorlinks,urlcolor=blue}
\usepackage{natbib}

\usepackage{xcolor}
\usepackage{graphicx}
\usepackage{graphics}
\usepackage{subcaption}
\usepackage{wrapfig}
\usepackage{enumitem}
\usepackage{verbatim}
\usepackage{float}
\usepackage[ruled,vlined]{algorithm2e}

\usepackage[colorinlistoftodos]{todonotes}

\usepackage[font=small,labelfont=bf]{caption}
\title{Differentiable Architecture Search for \\ Reinforcement Learning}

\newcommand\blfootnote[1]{%
  \begingroup
  \renewcommand\thefootnote{}\footnote{#1}%
  \addtocounter{footnote}{-1}%
  \endgroup
}

\author{%
  Yingjie Miao$^{*}$, Xingyou Song$^{*}$, John D. Co-Reyes, Daiyi Peng, Summer Yue, Eugene Brevdo, Aleksandra Faust \\
  Google Research, Brain Team\\
}

\newcount\Edits  %
\newcommand{\edit}[1]{\ifnum\Edits=1\textcolor{blue}{#1}\else{#1}\fi}

\begin{document}
\maketitle

\begin{abstract}
In this paper, we investigate the fundamental question: \textit{To what extent are gradient-based neural architecture search (NAS) techniques applicable to RL?} Using the original DARTS as a convenient baseline, we discover that the discrete architectures found can achieve up to 250\% performance \edit{compared to} manual architecture designs on both discrete and continuous action space environments across off-policy and on-policy RL algorithms, at only 3x more computation time. Furthermore, through numerous ablation studies, we systematically verify that not only does DARTS correctly upweight operations during its supernet phrase, but also gradually improves resulting discrete cells up to 30x more efficiently than random search, suggesting DARTS is surprisingly an effective tool for improving architectures in RL.
\end{abstract}

\blfootnote{$^\ast$Equal contribution. Order decided randomly.}
\blfootnote{Correspondence to: \texttt{\{yingjiemiao,xingyousong,sandrafaust\}@google.com}}
\blfootnote{Code can be found at \url{https://github.com/google/brain_autorl/tree/main/rl_darts}.}

\vspace{-0.6cm}
\section{Introduction and Motivation}
\label{sec:introduction_and_motivation}

\begin{wrapfigure}[13]{r}{0.5\textwidth}
\vspace{-17pt}
\begin{center}
    \includegraphics[keepaspectratio, width=0.5\textwidth]{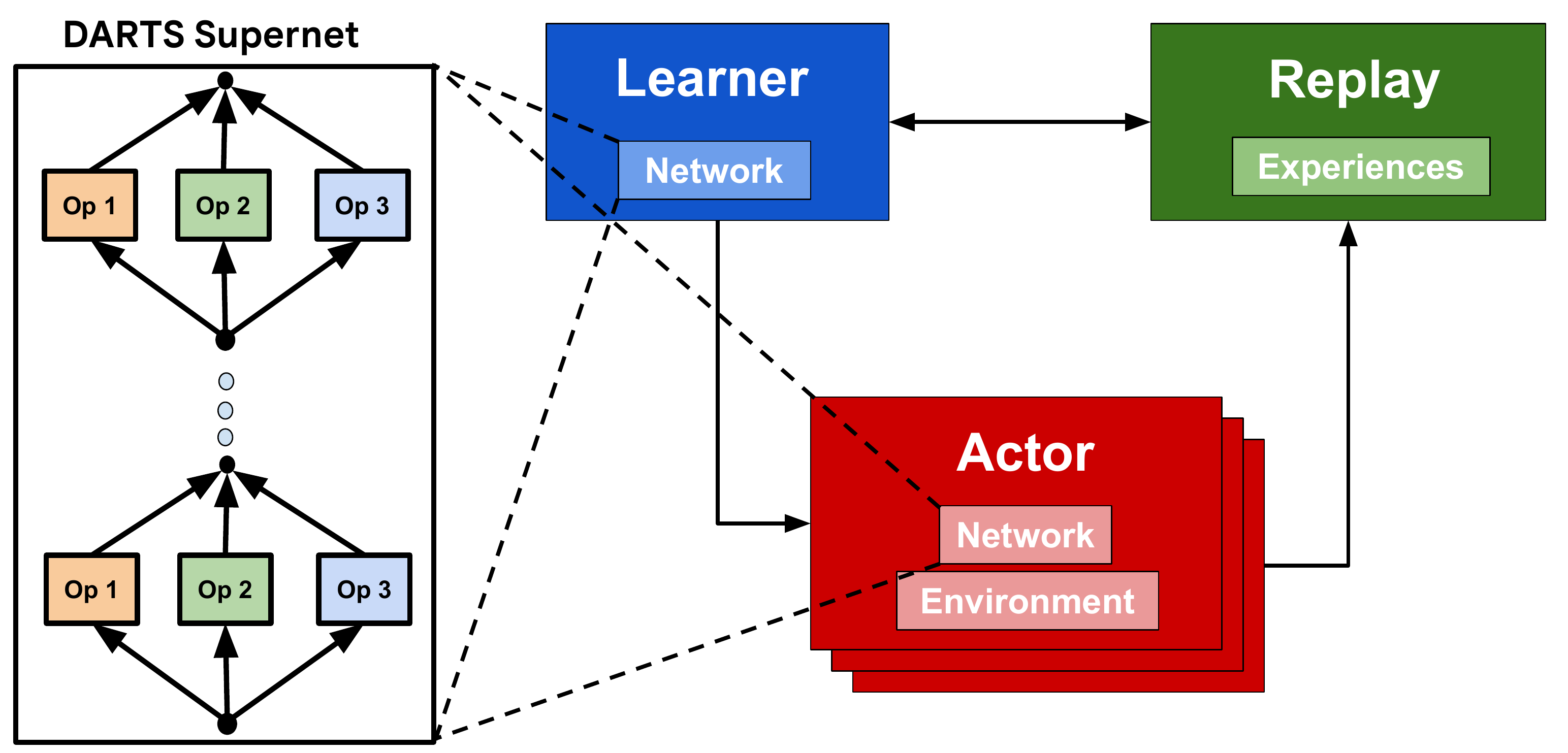}
\end{center}
\vspace{-13pt}
\caption{DARTS is potentially a natural fit for RL, as a DARTS supernet can simply be inserted into the network components of a standard RL training pipeline, which may potentially be highly distributed.}
\label{fig:rl_darts_pipeline}
\end{wrapfigure}

Over the past few years, Differentiable Architecture Search (DARTS) \citep{darts} has dramatically risen in popularity as a method for neural architecture search (NAS), with multiple modifications and improvements \citep{fair_darts, darts_geometry_aware, darts_minus, darts_plus, idarts, noisy_darts, sharp_darts, rethinking_darts, stabilizing_perturbation_darts, understanding_robustifying_darts} constantly proposed at a rapid speed. From a broader perspective, one may wonder how far we may push the limits of differentiable search, as differentiability is the cornerstone of all deep learning research. One such very large application space is in reinforcement learning (RL), with DARTS being a natural fit due to its simplicitly and modularity in integrating with large distributed RL training pipelines. Shockingly however, among the vast amounts of literature on DARTS, there are virtually no works addressing its viability in RL.

This can be attributed to the fact that RL fundamentally does not follow the same optimization paradigm as \edit{supervised learning (SL).} The end goal of RL is not to simply minimize the loss or accuracy over a fixed dataset, but rather to improve a policy's reward over an entire environment whose training data is generated by the policy itself. This scenario raises the possibility of a negative feedback loop in which a poorly trained policy may achieve trivial reward even if it successfully optimizes its loss to zero, and thus the loss is not an indicator of a policy's true performance. As DARTS is purely reliant on optimizing the architecture topology by using only the loss as the search signal, therein lies the simple question: \textit{Would DARTS even work in RL?}

It is highly valuable to answer this question, as recently, works in larger-scale RL suggest that policy architecture designs greatly affect various metrics such as generalization, transferrability, and efficiency. One surprising phenomenon found on Procgen \citep{procgen}, a procedural generation benchmark for RL, was that "IMPALA-CNN", a residual convolutional architecture from \citep{impala_paper}, could substantially outperform "NatureCNN", the standard 3-layer architecture used for Atari \citep{atari}, in both generalization and sample complexity under limited and infinite data regimes respectively \citep{coinrun}. Furthermore, in robotics subfields such as grasping \citep{qt_opt, rl_cyclegan}, cameras collect very detailed real-world images (ex: 472 x 472, 4x larger than ImageNet \citep{imagenet}) for observations which require deep image encoders consisting of more than 15 convolutions, raising concerns on efficiency and speed in policy training and inference. As such RL policy networks gradually become larger and more sophisticated, so does the need for understanding and automating such designs.

In this paper, we show that DARTS can in fact find better architectures efficiently in RL, one of the first instances in which the loss does not directly affect the final objective. This work may be of interest to both the gradient-based NAS community as means to expand to the RL domain, and the AutoRL community as a practical toolset and method for co-training policies and neural architectures for better performing agents. Our contributions are:

\begin{itemize}
\item We conceptually identify the key differences between SL and RL in terms of their usage of the loss function, which raise important hypothetical questions and issues about whether DARTS is applicable to RL. In particular, these deal with the quality of the training signal to the architecture variables during supernet training, and downstream effects on discrete cells during evaluation.

\item Empirically, we find that DARTS is in fact compatible with several on-policy and off-policy algorithms including PPO \citep{ppo}, Rainbow-DQN \citep{rainbow_dqn}, and SAC \citep{sac}. The discrete architectures found can reach up to \edit{250\% performance compared to} manual architecture designs on discrete action (e.g. Procgen) and continuous control (e.g. DM-Control) environments, at only 3x more computation time.

\item Through comprehensive ablation studies, we show the supernet successfully trains, and reasonably upweights optimal operations. We further verify both qualitatively and quantitatively that discretized cells gradually evolve to better architectures. However, we also demonstrate how this can fail, especially if the corresponding supernet fails to train, with further extensive ablations in the Appendix.
\end{itemize}

\paragraph{Related Works} \edit{Recently, there have been a flurry of works modifying many components in the RL pipeline, both manually and automatically, as part of the broader Automated Reinforcement Learning (AutoRL) \citep{autorl_survey} field. Specifically for architecture components, manual modifications include \citep{decoupling_value_policy, attention_decoupling, sensory_neuron_transformer} which have shown great success in improving metrics such as sample complexity and generalization, especially on the Procgen benchmark. However, very few works have considered the possibility of actually \textit{automating} the search for new architectures, i.e. "NAS for RL", specifically for large-scale modern convolutional networks.}

Most previous NAS for RL works only involve small policies trained via blackbox/evolutionary optimization methods, which include \citep{es_enas, wann, rl_neat, hyperneat}, utilizing CPU workers for forward pass evaluations rather than exact gradient computation on GPUs. Such methods are usually unable to train policies involving more than 10K+ parameters due to the sample complexity of zeroth order methods in high dimensional parameter space \citep{es_sample_complexity}. The only previous known application of gradient-based routing is \citep{visionary}, which searches for the optimal way of combining observation and action tensors together in off-policy QT-OPT \citep{qt_opt}, but does not search for image encoders nor uses the supernet for inference, as it trains using off-policy robotic data collected independently. This leaves the applicability of DARTS to inference-dependent RL as an open question addressed in our work.

\section{Problem Overview and Method}
\label{sec:rl_darts_method}

\paragraph{DARTS Preliminaries} Since we only use the original DARTS \citep{darts} to reduce confounding factors, we thus give a very brief overview of DARTS to save space. More comprehensive details can be found in Appendix \ref{subsec:training_procedure} and the original paper. \edit{DARTS optimizes substructures called cells, where each cell contains $I$ intermediate nodes organized in a directed acyclic graph, where each node $x^{(i)}$, represents a feature map, and each edge $(i,j)$ consists of an operation (op) $o^{(i,j)}$, with later nodes $x^{(j)}$ merged (e.g. summation) from some previous $o^{(i,j)}(x^{(i)})$. A DARTS supernet is constructed by continuously relaxing selection of ops in $\mathcal{O}$, via softmax weighting, i.e. $\overline{o}^{(i,j)}(x^{(i)}) = \sum_{o \in \mathcal{O}} p_{o}^{(i,j)} \cdot o(x^{(i)})$, where $p_{o}^{(i,j)}=  \frac{\exp(\alpha_{o}^{(i,j)})}{ \sum_{o' \in \mathcal{O}} \exp(\alpha_{o'}^{(i,j)}) }$. The cell's output is by default the result of a Conv1x1 op on the depthwise concatenation of all intermediate node features, although this may be changed (e.g. by simply outputting the last intermediate node's features). We denote the collection of all architecture variables $a_{o}^{(i,j)}$ as $\alpha$.} Denote the total set of possible operations in our searchable network as $\mathcal{O} = \mathcal{O}_{base} \cup \{\text{Zero, Skip}\}$ which must contain Zero and Skip Connection ops, while $\mathcal{O}_{base}$ is user-defined. Denote $\alpha$ to be the pre-softmax trainable architecture variables in the supernet. A predefined loss function $\mathcal{L}(\theta)$ over neural network weights $\theta$ will thus be redefined as $\mathcal{L}(\theta, \alpha)$ when under DARTS's \textit{search mode}, where the original model $f_{\theta}$ will be replaced with a dense supernet $f_{\theta, \alpha}$. During evaluation time, a trained $\alpha^{*}$ is then \textit{discretized} into a sparser final cell $\delta(\alpha^{*})$ \edit{by representing each edge $(i,j)$ with the highest softmax weighted op, i.e. $\arg \max_{o \in \mathcal{O}, o \neq zero} p_{o}^{(i,j)}$, and then retaining only the top $K$ incoming edges for each intermediate node.} We thus retrain \edit{over} the new loss $\mathcal{L}_{\delta(\alpha^{*})}(\phi)$, now dependent on only fresh sparse weights $\phi$ to obtain the final reported metric.

\paragraph{RL Preliminaries} For RL notation, given an MDP $\mathcal{M}$, denote $s_{t}, a_{t}, r_{t}$ as state, action, reward respectively at time $t$. $\pi$ is the policy and $\mathcal{D}$ is the replay buffer containing collected trajectories $\tau = (s_{0}, a_{0}, r_{0}, s_{1},\ldots)$. The goal is to maximize $J(\pi) = \mathbb{E}_{\tau \sim \pi}\left[\sum_{t \ge 0} r_{t}\right]$, the expected cumulative reward using policy $\pi$. In most RL algorithms, there is the notion of a neural network torso or \textit{encoder} $f_{\theta}$ mapping the state $s$ to a final feature vector when forming $\pi$. In the DARTS case, we use a supernet encoder $f_{\theta, \alpha}$ leading to a supernet policy denoted as $\pi_{\theta, \alpha}$ and also a corresponding discretized-cell policy $\pi_{\phi, \delta(\alpha^{*})}$ for evaluation.

\begin{algorithm}[t]
\SetAlgoLined
\textbf{1. Supernet training:} Compute $\alpha^{*}$ from $\argmax_{\theta, \alpha} J(\pi_{\theta, \alpha})$ via $\argmin_{\theta, \alpha} \mathcal{L}^{RL}(\theta, \alpha)$. \\
\textbf{2. Discretization:} Discretize $\alpha^{*}$ to construct evaluation policy $\pi_{\phi, \delta(\alpha^{*})}$. \\
\textbf{3. Evaluation:} Report $\max_{\phi} J(\pi_{\phi, \delta(\alpha^{*})})$ via $\argmin_{\phi} \mathcal{L}_{\delta(\alpha^{*})}(\phi).$
\caption{RL-DARTS Procedure.}
\label{algo:rl_darts}
\end{algorithm}

\subsection{Methodology}
\label{subsec:sl_vs_rl}
SL features the notion of training and validation sets, with corresponding losses $\mathcal{L}^{SL}_{train}, \mathcal{L}^{SL}_{val}$, where the learning procedure consists of a bilevel optimization problem and the goal is to find $\alpha^{*} = \argmin_{\alpha} \mathcal{L}^{SL}_{val}(\theta^{*}, \alpha)$ where $\theta^{*} = \argmin_{\theta} \mathcal{L}^{SL}_{ train}(\theta, \alpha)$. In this paper, we do not need to use the original bilevel optimization framework, as we are optimizing sample complexity and raw training performance, which are standard metrics in RL. Furthermore, bilevel optimization is notoriously difficult and unstable, sometimes requiring special techniques \citep{progressive, gdas, sharp_darts, sgas, noisy_darts, darts_minus, fair_darts, darts_plus, rethinking_darts} specific to SL optimization, which can confound the results and message of our paper. 

We thus joint optimize both $\theta$ and $\alpha$, and the full procedure is concisely summarized in Algorithm \ref{algo:rl_darts} as "RL-DARTS". However, this is easier said than done, as we explain the core issues of applying DARTS to RL below.

\paragraph{SL vs RL DARTS} Fundamentally, SL relies on a fixed dataset $\mathcal{D}^{SL} = \{(x_{i}, y_{i}) \> \> | \> \> i \ge 1 \}$, in which the loss is defined as $\mathcal{L}^{SL}(\theta) = \mathbb{E}_{(x,y) \sim \mathcal{D}} \left[ \ell(f_{\theta}(x), y) \right]$ where $\ell(\cdot)$ is defined as mean squared error, cross-entropy loss, or negative log-likelihood depending on application. These losses are strongly correlated or even equivalent to the final objective (e.g. accuracy or density estimation), and this reason can be considered a significantly contributing factor to the success of DARTS in SL. Unfortunately in RL, there are no such guarantees that minimizing the loss $\mathcal{L}^{RL}(\cdot)$ necessarily improves the true objective $J(\cdot)$, for two primary reasons:

\begin{enumerate}

\item The RL agent's dataset (a.k.a. replay buffer) $\mathcal{D}^{RL} = \{ \tau_{i} \> \> | \> \> i \ge 1\}$ is significantly non-stationary and self-dependent, as it constantly changes based on the current performance of data collection actors, which themselves are functions of $\theta$. Thus, a negative feedback loop may arise, where $\theta$ produces an actor which collects poor training data, leading to convergence towards an even poorer $\theta'$. While the loss $\mathcal{L}^{RL}(\cdot)$ converges to 0 over low quality data, the reward $J(\cdot)$ still does not increase. This issue is commonplace in RL, such as in any environments which require exploration. Hypothetically in the DARTS case, $\alpha$ can potentially produce the same negative feedback loop by converging to subpar operations and impairing supernet training, also leading to a poor discrete cell $\delta(\alpha)$.

\item The losses are considerably more complex and utilize multiple auxiliary losses which are used to assist training but never used during evaluation. For PPO, the loss is defined as $\mathcal{L}^{RL}_{PPO}(\theta) = \mathbb{E}_{\tau \sim \mathcal{D}_{RL}} \left[L_{CLIP}(\theta) - L_{VF}(\theta) + \mathcal{H}(\theta) \right]$. However, the value function (corresponding to $L_{VF}$) nor the policy entropy (corresponding to $\mathcal{H}$) are never used to evaluate final reward. This is similarly the case for off-policy algorithms such as SAC which strongly emphasizes maximizing the entropy $\mathcal{H}$ and Rainbow-DQN which also uses value functions to assist training. The question is thus raised in the DARTS case as to whether such auxiliary losses are actually useful signals, or inhibit proper training of $\alpha$, whose main goal is only maximize $J(\cdot)$ using discrete cell $\delta(\alpha)$.
\end{enumerate}

The core theme of our experiments will be understanding whether these are obstacles to DARTS's application to RL.

\section{Experiments}
\label{sec:experiments}
\textbf{Experiment Setup:} To verify that every component of RL-DARTS works as intended, we seek to answer all questions below, by first presenting end-to-end results, and then further key ablation studies:
\begin{enumerate}[itemsep=-0.4mm]
\item \textbf{End-to-End Performance:} Overall, how do the final discrete cells perform at evaluation, and what gains can we obtain from architecture search? Furthermore, how does RL-DARTS compare against random search? 
\item \textbf{Supernet Training:} During supernet training, how does the $\alpha$ change? Does $\alpha$ converge towards a sparse solution and select good operations over supernet training?
\item \textbf{Discrete Cells:} Even if the supernet trains, do the corresponding discrete cells also improve in evaluation performance throughout $\alpha$'s training? What kinds of failure modes occur?
\end{enumerate}

\begin{wrapfigure}[17]{r}{0.5\textwidth}
\vspace{-12pt}
\begin{center}
    \includegraphics[keepaspectratio, width=0.475\textwidth]{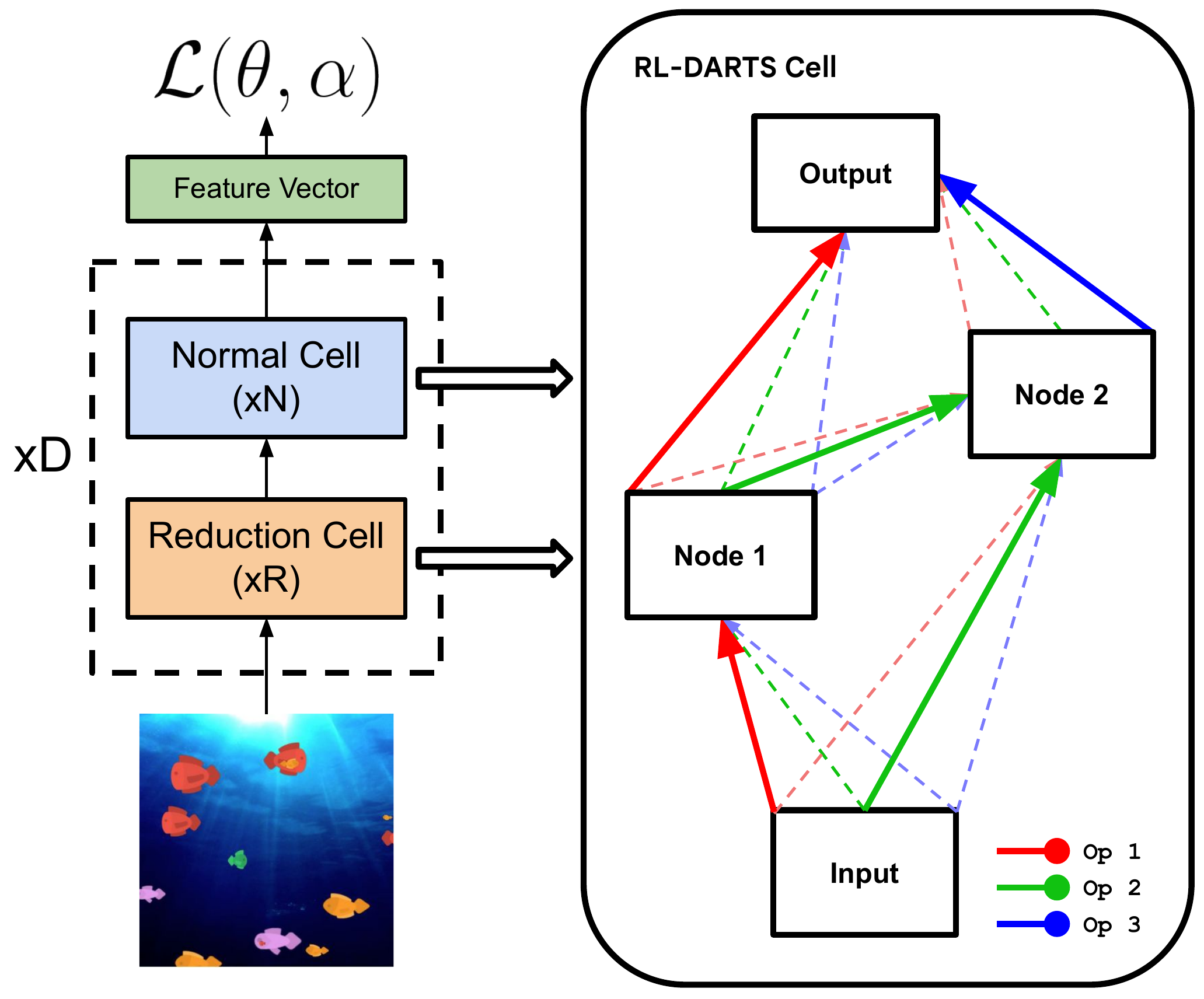}
\end{center}
\vspace{-17pt}
\caption{Illustration of our network via stacking normal and reduction cells. Solid lines correspond to selected ops after discretization from all possible ops weighted using $\alpha$. If $R>0$, we add an initial Conv3x3 for preprocessing.}
\label{fig:rl_darts}
\end{wrapfigure}

Following common NAS practices \citep{image_recognition, enas, original_nas}, we construct our supernet (with $I$ intermediate cells) by stacking both normal ($N$ times) and reduction cells ($R$ times) together into \textit{blocks}, which are themselves also stacked together $D$ times (see Figure \ref{fig:rl_darts}). Reduction cells apply a stride of 2 on the input. Each block possesses its own convolutional channel depth, used throughout all cells in the block. During search, we train a smaller supernet (i.e. depth 16) to reduce computation time, but evaluate final discretized cells on larger models (i.e. depth 64), with $D =3$ layers for cheap large-scale runs and $D=5$ layers for fine-grained A/B testing.

For the operation search space, we consider the following base ops $\mathcal{O}_{base}$ and corresponding algorithms:

\begin{itemize}
\item \textbf{Classic on PPO:} $\mathcal{O}_{base, N}$ = \{Conv3x3+ReLU, Conv5x5+ReLU, Dilated3x3+ReLU, Dilated5x5+ReLU\} for normal ops and $\mathcal{O}_{base, R}$ = \{Conv3x3, MaxPool3x3, AveragePool3x3\} for reduction ops, which is standard in supervised learning \citep{darts, image_recognition, enas}. 

\item \textbf{Micro on Rainbow and SAC:} We also propose $\mathcal{O}_{base, N}$ = \{Conv3x3, ReLU, Tanh\}, a more fine-grained and novel search space which has not been used previously in SL. The inclusion of Tanh is motivated by its use previously for continuous control architectures \citep{es,observational}. 
\end{itemize}

For benchmarks, we primarily use Procgen \citep{coinrun, procgen} for discrete action spaces, with PPO \citep{ppo} and Rainbow DQN \citep{rainbow_dqn} as training algorithms, with Procgen's difficulty set to "easy" similar to other works \citep{autodrac, decoupling_value_policy, tuning_mixed_hps}. Procgen comprehensively evaluates all aspects of RL-DARTS, as it possesses a diverse selection of 16 games, each with infinite levels to simulate large data regimes where episodes may drastically change, relevant for generalization. It further uses the IMPALA-CNN architecture \citep{impala_paper} as a strong hand-designed baseline, and can be seen as a specific instance of the stacked cell design in Figure \ref{fig:rl_darts}, where its "Reduction Cell" consists of a Conv3x3 and MaxPool3x3 (Stride 2) with $R=1$ and its "Normal Cell" consists of a residual layer with Conv3x3's and ReLU's, with $N=2$. For fair comparisons to IMPALA-CNN, we use $(N,R,I) = (1,1,4)$ on the classic search space, while $(N,R,I) = (2,0,4)$ on the Micro search space, where reduction ops default to IMPALA-CNN's in order to avoid hidden confounding effects when visualizing Micro cells.

In addition, we also assess DARTS's viability in continuous control which is common in robotics tasks. We use the common DM-Control benchmark \citep{dm_control} along with the popular SAC algorithm. We use $N=3, I=4, K=1$ with "Micro" search space to remain fair to the 4-layer convolutional encoder baseline observing images of size $64 \times 64$ (full details in Appendix \ref{appendix:hyperparameters}).

Unless specified, we by default use consistent hyperparameters (found in Appendix \ref{appendix:hyperparameters}) for all comparisons found inside a figure, although learning rate and minibatch size may be altered when training models of different sizes due to GPU memory limits. Thus, even though RL is commonly sensitive to hyperparameters \citep{model_rl_hp_tuning}, we surprisingly find that \textbf{once a pre-existing RL baseline has already been setup, incorporating DARTS requires no extra cost in tuning, as evidence of its ease-of-use.}

\subsection{End-to-End Results on Multi-task, Discrete and Continuous Control Tasks}

\begin{figure}[h]
    \centering
    \setlength\tabcolsep{4pt}    
    \begin{tabular}{c c}
        \smallskip
        \begin{subfigure}{0.49\textwidth}
            \centering
            \includegraphics[keepaspectratio, width=1.0\textwidth]{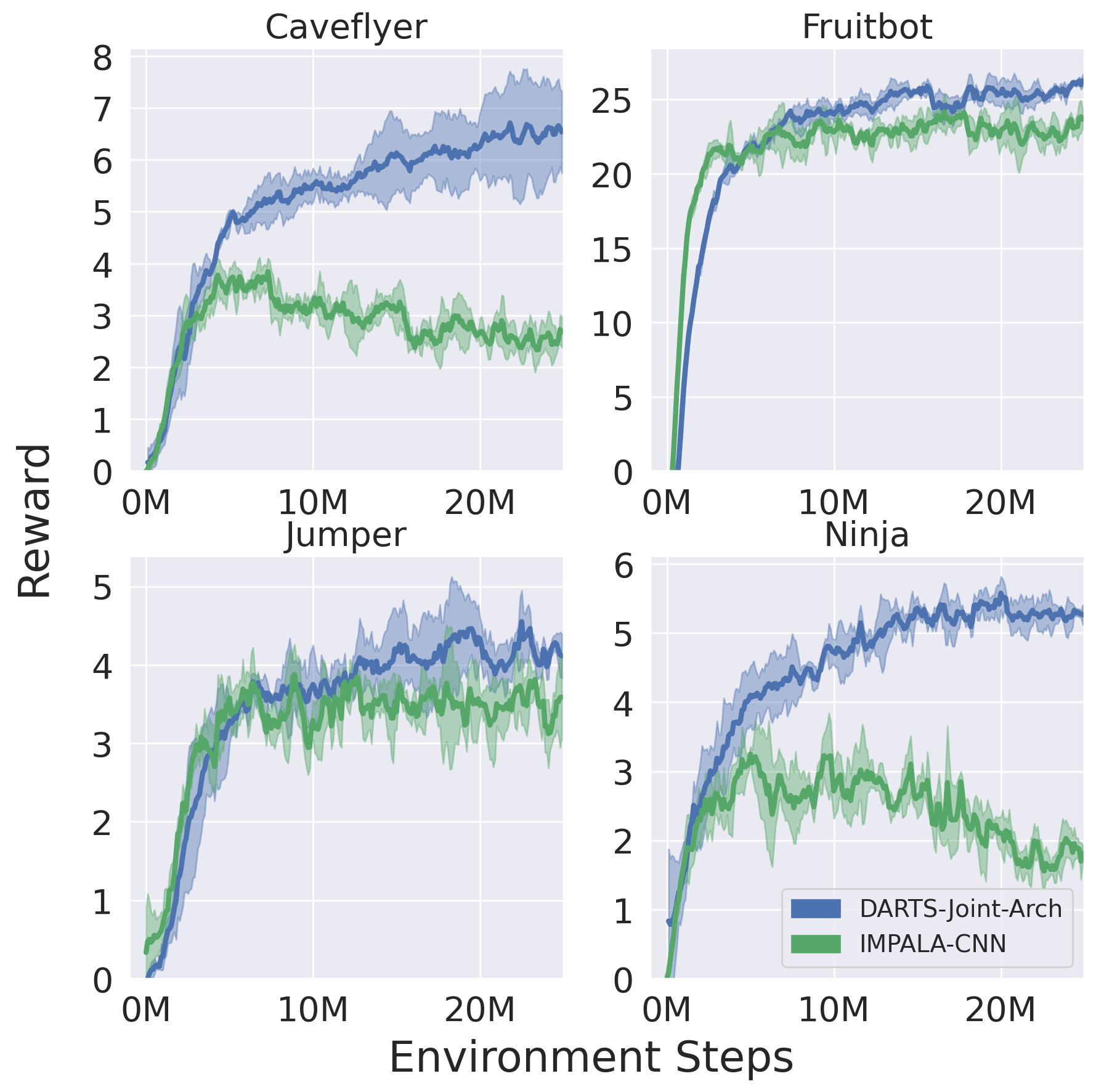}
        \end{subfigure}
             &  
        \begin{subfigure}{0.49\textwidth}
        \centering
        \includegraphics[keepaspectratio, width=1.0\textwidth]{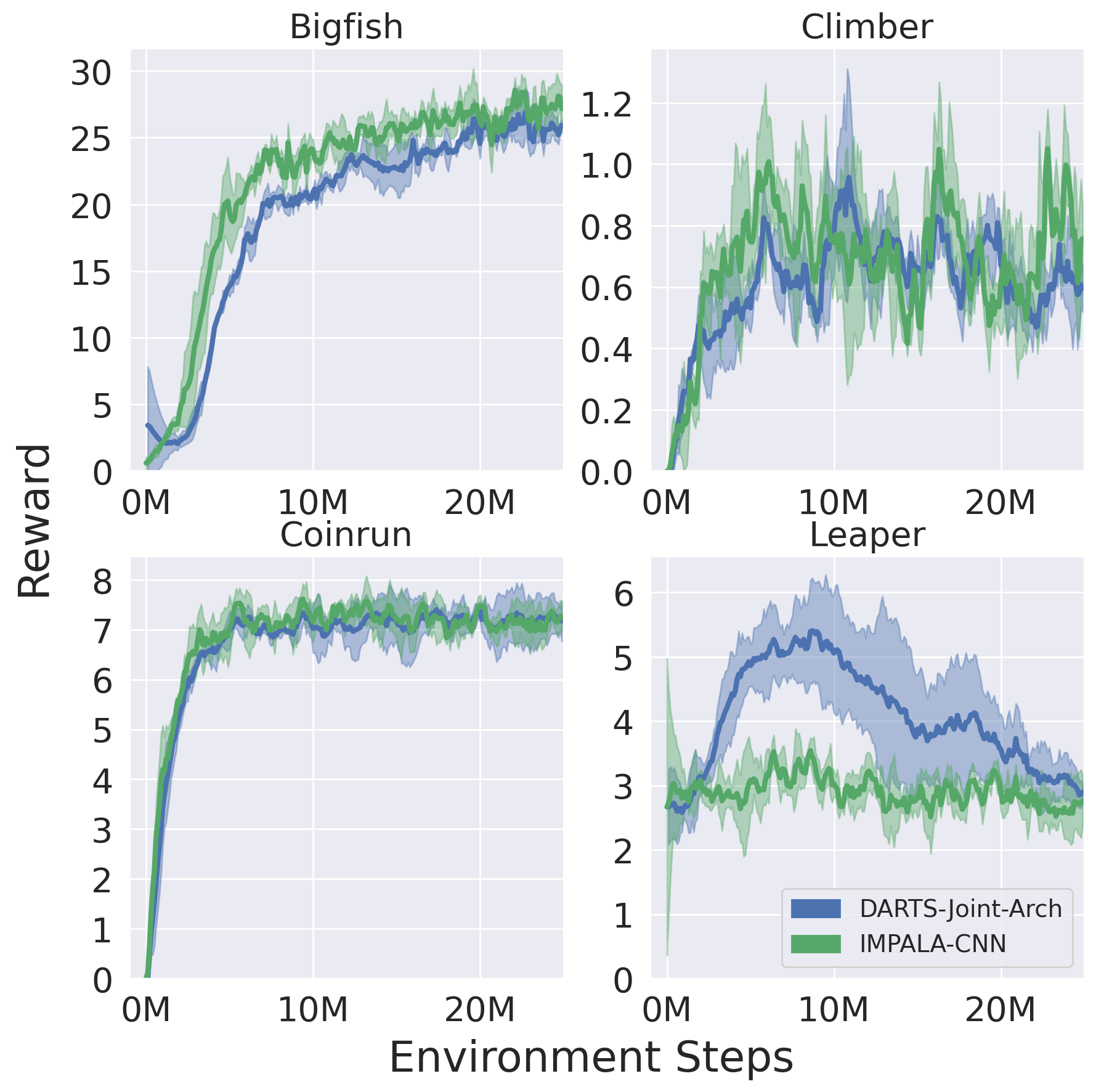}
        \end{subfigure}
    \end{tabular}
    \vspace{-5pt}
    \caption{Evaluation of the discrete cell joint-trained over 8 environments using depths $64 \times 5$ to emphasize comparison differences.}
    \label{rainbow_micro_eval_transfer}
\end{figure}

\paragraph{Multi-game Search} We first begin with the most surprising and largest end-to-end result in terms of scale of data and compute shown in Figure \ref{rainbow_micro_eval_transfer}: \textbf{By training a supernet across infinite levels across \textit{all 16 Procgen games} to find a single transferrable cell, we are able to achieve up to \bm{$250\%$} \edit{evaluation performance compared to} the IMPALA-CNN baseline} over select environments. This is achieved using Rainbow with our proposed "Micro" search space, where a learner performs gradient updates over actor replay data (with normalized rewards) from all games.

\label{subsec:end_to_end}

\begin{figure}[h]
    \centering
    \begin{subfigure}{0.49\textwidth}
    \includegraphics[keepaspectratio, width=1.0\textwidth]{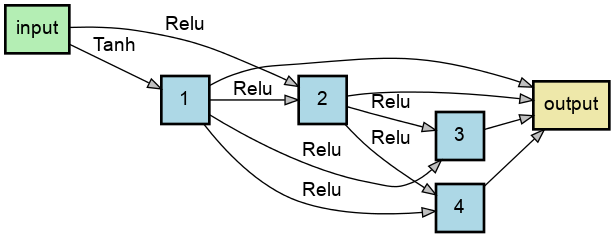}
    \end{subfigure}
    \begin{subfigure}{0.49\textwidth}
    \includegraphics[keepaspectratio, width=1.0\textwidth]{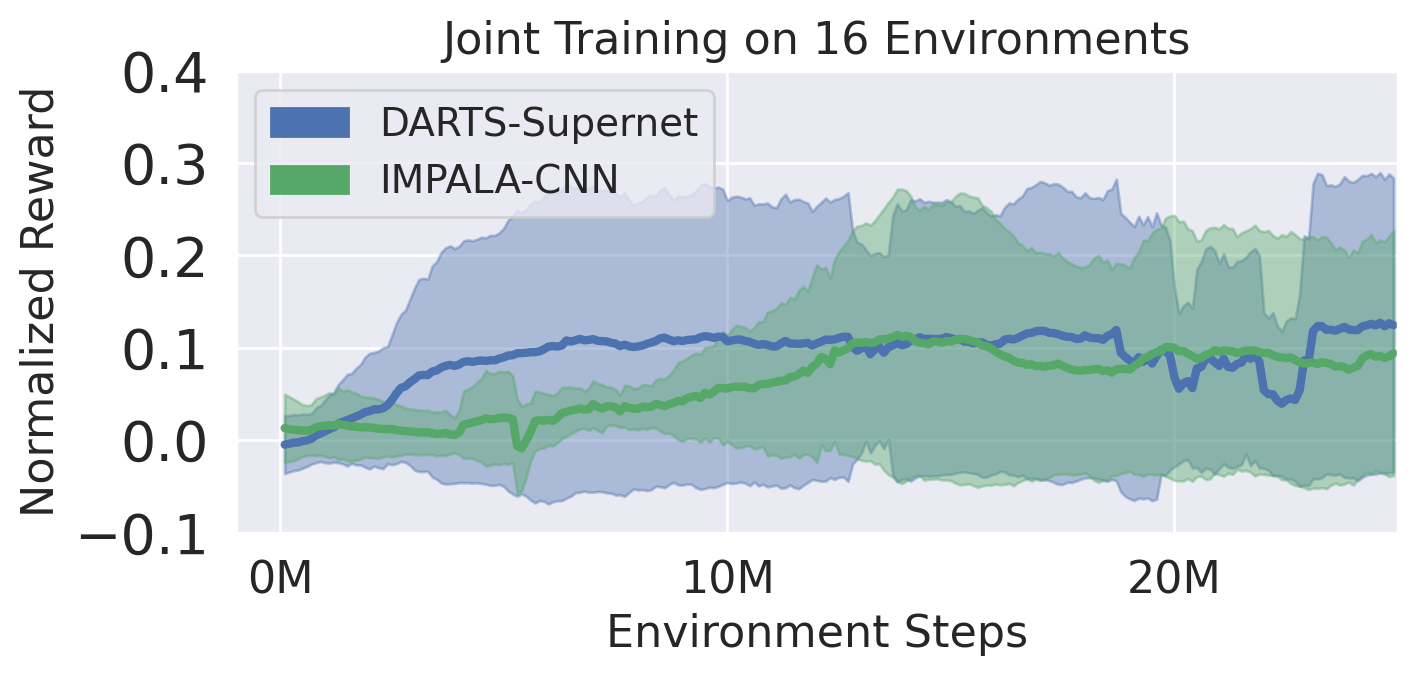}
    \end{subfigure}
    \caption{\textbf{Left:} Discrete cell found. \textbf{Right:} Average normalized rewards over all 16 games during supernet + baseline training.}
    \label{rainbow_micro_eval_transfer_aux}
\end{figure}

\edit{We further display the discovered discrete cell and the supernet joint-training procedure in Figure \ref{rainbow_micro_eval_transfer_aux}. Interestingly, the discrete cell uses nonlinearities over all intermediate connections, with convolutions only used via the merge operation for the output.}
 
\paragraph{Single-game Search} We further compare DARTS's end-to-end performance against random search, but more appropriately applied to single game scenarios. On the PPO side, we use the "Classic" search space (total size $4 \times 10^{11}$, see Appendix \ref{appendix:search_space_size}). For a fair comparison, we ensure total wall-clock time (with same hardware) stays equal, as common in \citep{darts}. Since in Appendix \ref{appendix:efficiency}, Table \ref{table:efficiency}, a PPO supernet takes 2.5x longer to reach 25M steps, this is rounded to a random search budget of 3 cells to be trained with depths $16 \times 3$ for 25M steps. The best of the 3 cells is used for full evaluation. In Table \ref{table:ppo_normalized_summary}, on average, random search underperforms significantly.

\vspace{0.2cm}
\begin{table}[h]
\centering
\begin{tabular}{cccc}
\hline
& IMPALA-CNN & RL-DARTS (Discrete Cell) & Random Search \\
\hline
Avg. Normalized Reward & 0.708 & 0.709 & 0.489 \\
\hline
\end{tabular}
\caption{Average normalized rewards across all 16 environments w/ PPO, using the normalization method from \citep{procgen}. Full details and results (including Rainbow) are presented in Appendix \ref{sec:numerical}.}
\label{table:ppo_normalized_summary}
\end{table}

\begin{figure}[h]
    \centering
    \setlength\tabcolsep{4pt}    
    \begin{tabular}{c c}
        \smallskip
        \begin{subfigure}{0.49\textwidth}
            \centering
            \includegraphics[keepaspectratio, width=1.0\textwidth]{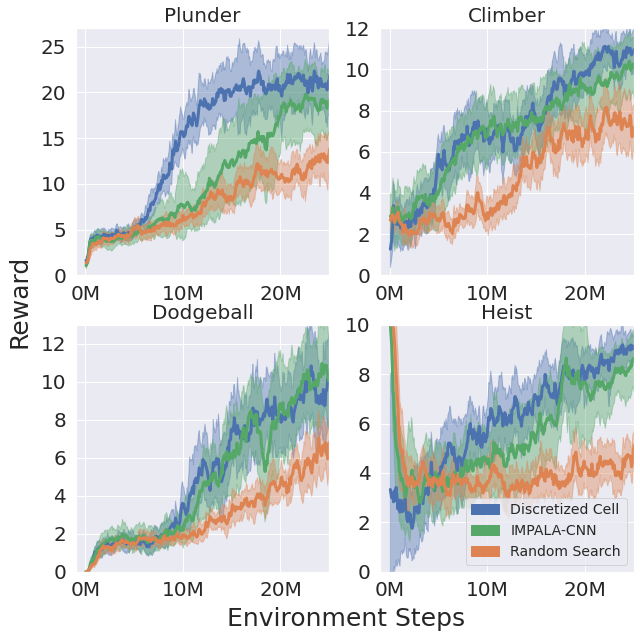}
        \end{subfigure}
        \begin{subfigure}{0.49\textwidth}
        \centering
        \includegraphics[keepaspectratio, width=1.0\textwidth]{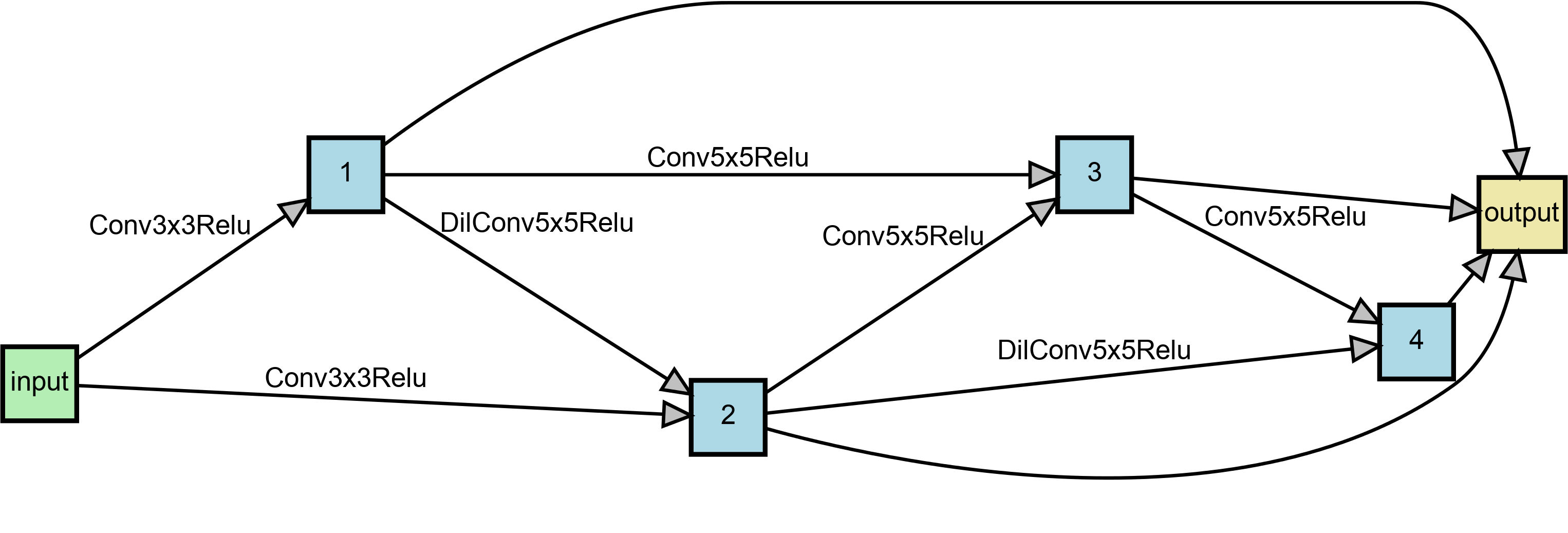}
        \includegraphics[keepaspectratio, width=1.0\textwidth]{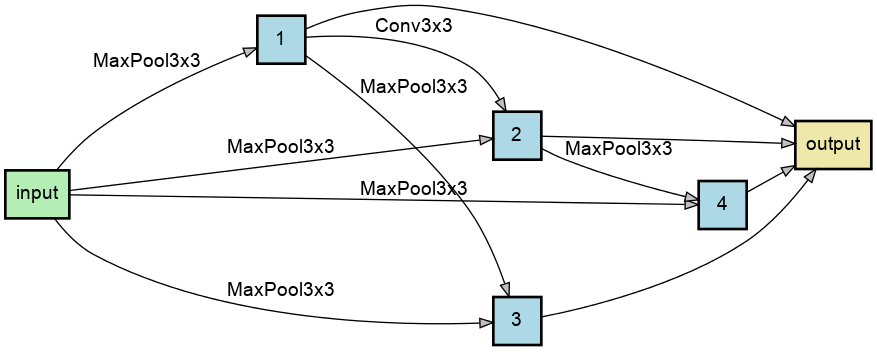}
        \end{subfigure}
    \end{tabular}
    \caption{\textbf{Left 2x2 Plot:} Examples of discrete cell evaluations using the "Classic" search space with PPO, with depths $64 \times 3$. \textbf{Right:} Normal (Top) and Reduction (Bottom) cells found for "Plunder" which achieves faster training than IMPALA-CNN. Note the interesting use of 5x5 convolutional kernel sizes later in the cell.}
    \label{ppo_clasic_eval}
\end{figure}

In Figure \ref{ppo_clasic_eval} we further find that RL-DARTS is capable of finding game-specific architectures from scratch which outperform IMPALA-CNN on select environments such as Plunder and Heist, while maintaining competitive performance on others.
 
\begin{figure}[h]
    \centering
    \includegraphics[keepaspectratio, width=0.49\linewidth]{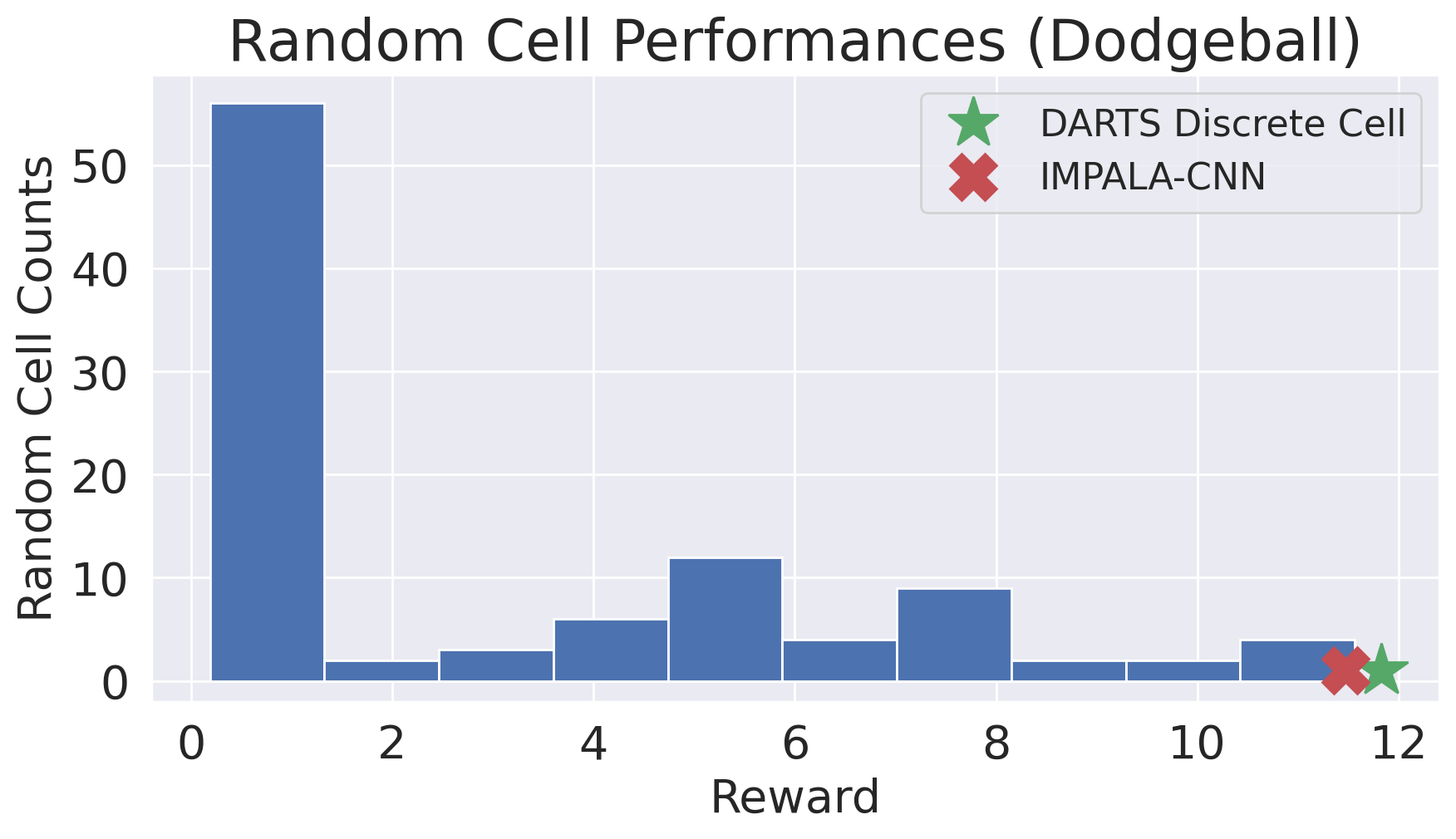}
    \includegraphics[keepaspectratio, width=0.49\linewidth]{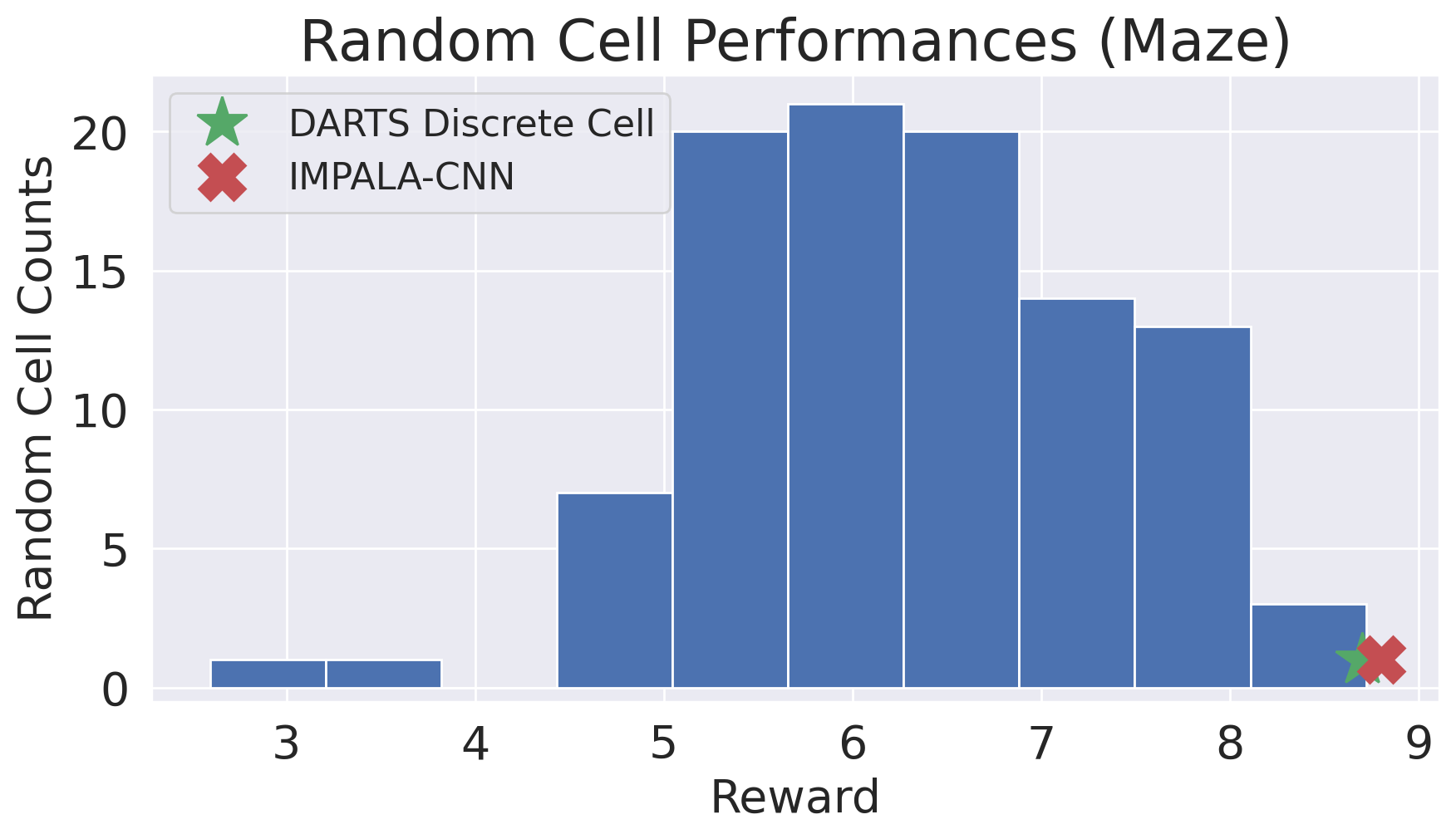}
    \caption{Histogram of 100 random cells' rewards over environments Dodgeball and Maze using the Rainbow + "Micro" search space (depths $64 \times 3$), \edit{with a significant number of random cells (e.g. 95\% for Dodgeball) performing substantially worse than DARTS or IMPALA-CNN.}}
    \label{fig:games-random-cell-hist}
\end{figure}

\paragraph{100 Random Cell Comparison} To further present a more comprehensive comparison against random search, \edit{and understand how strong IMPALA-CNN is as a baseline,} we compare against \underline{100} unique random cells. \edit{To manage the computational load, we performed the study over two selected environments, as shown in Figure \ref{fig:games-random-cell-hist} and find that \textbf{all of the random cells underperform against both IMPALA-CNN and DARTS}}. \edit{This demonstrates that} DARTS possesses a strong search capability, achieving $100 / 3 \approx 30x$ efficiency over random search \edit{and can discover architectures that match in complexity and performance of highly-tuned, expert designed architectures such as IMPALA-CNN.}

\paragraph{DM-Control with Soft Actor-Critic} DARTS also consistently finds better and stable architectures over multiple lightweight environments involving continuous control (see Figure \ref{fig:sac}) trained up to 1M steps. \edit{Even though the 4-layer encoder network used (details in Appendix \ref{appendix:hyperparameters}) is significantly smaller than IMPALA-CNN, we find that there is still room for architectural improvement.}

\begin{figure}[h]
    \center
    \includegraphics[keepaspectratio, width=0.95\textwidth]{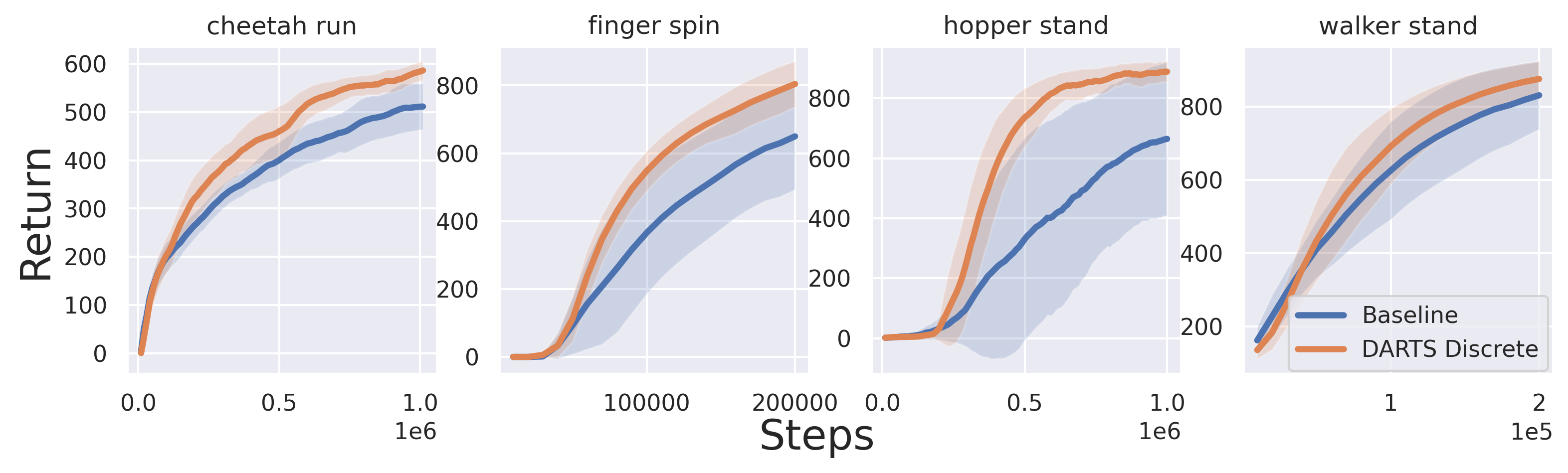}
\caption{RL-DARTS on DM-Control with SAC also finds better architectures over the corresponding baseline.}
\label{fig:sac}
\end{figure}

\newpage

\subsection{Role of Supernet Training}
\label{subsec:supernet_training}
\begin{figure}[h]
    \includegraphics[keepaspectratio, width=0.49\textwidth]{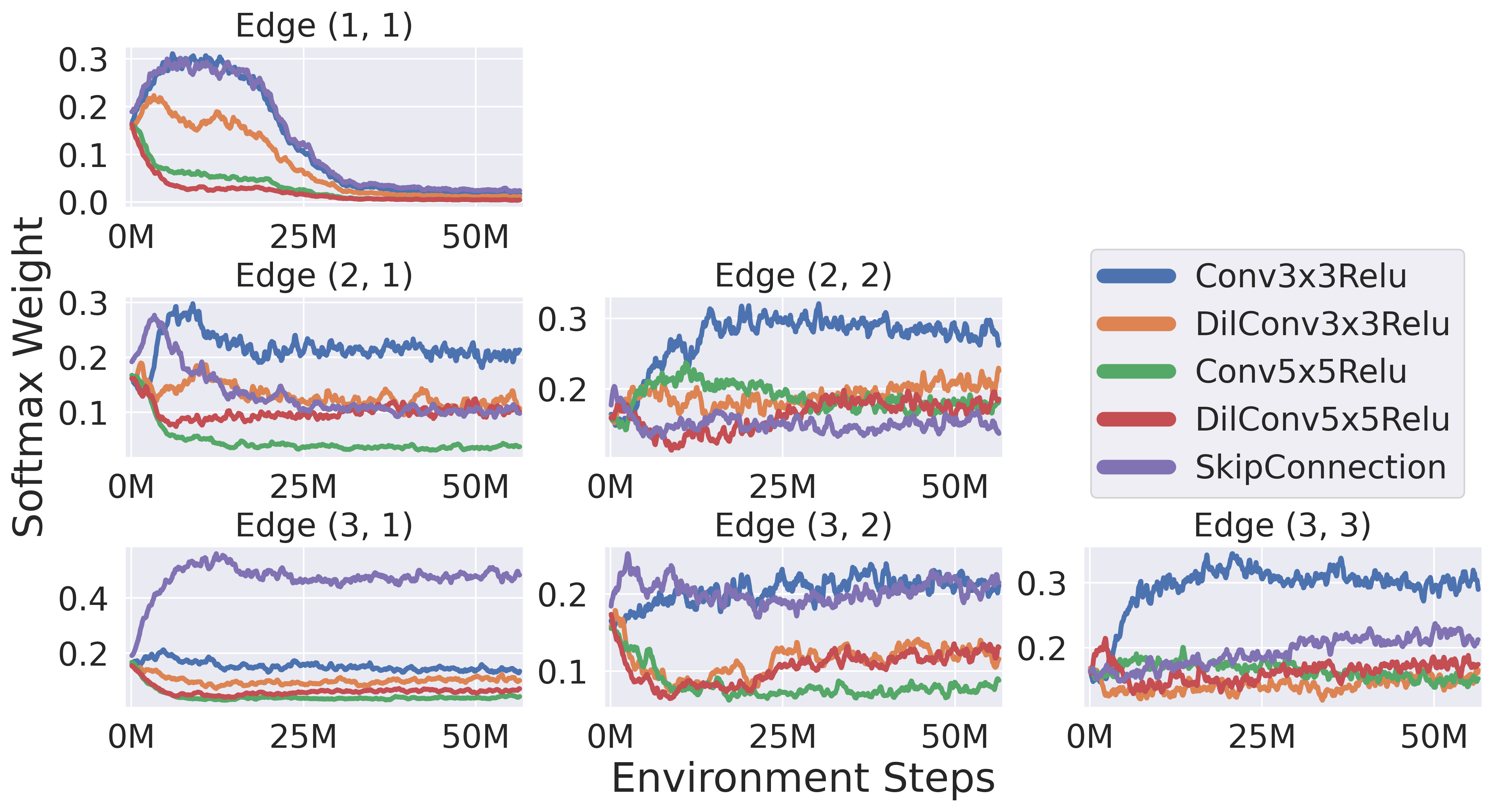}
    \includegraphics[keepaspectratio, width=0.5\textwidth]{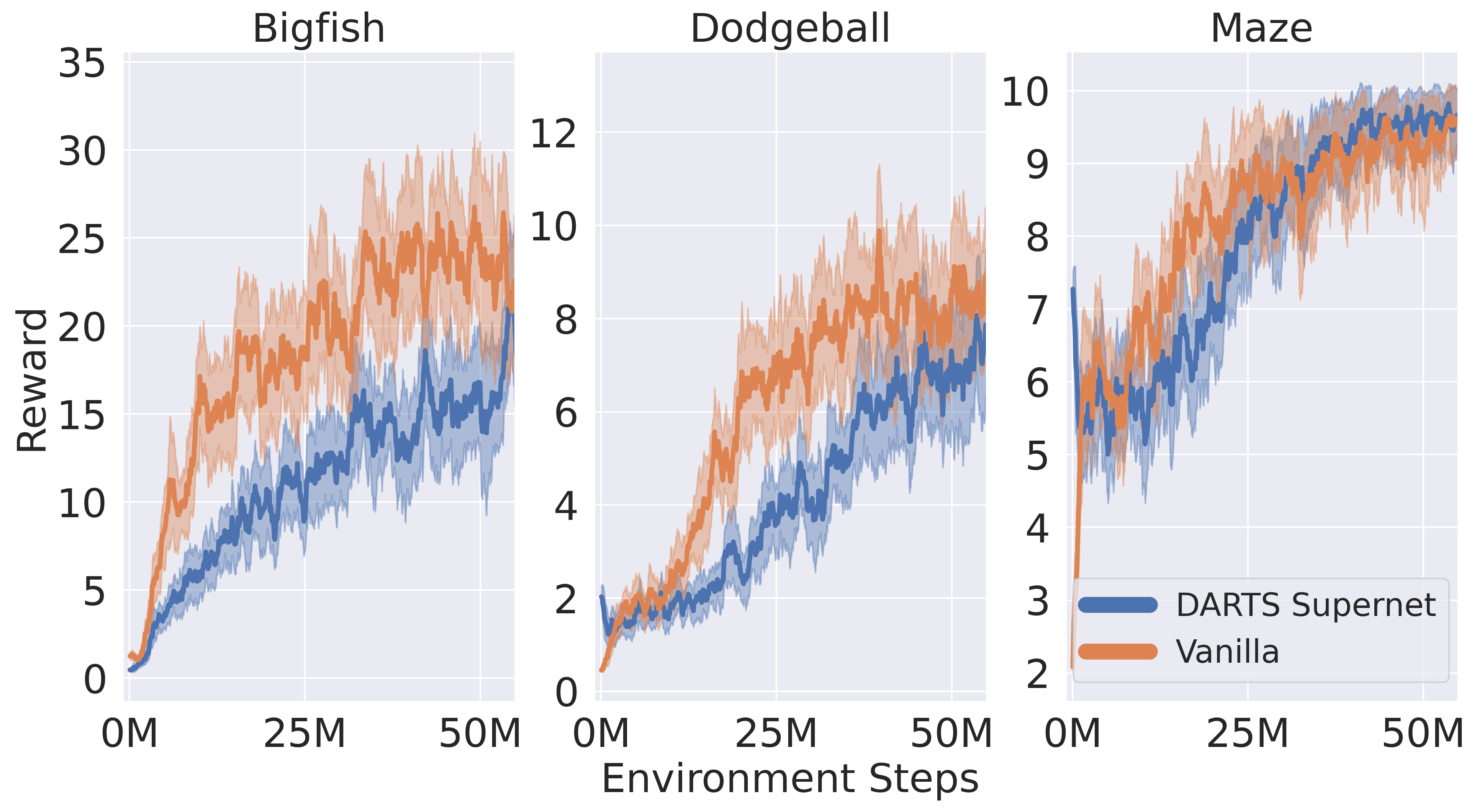}
\caption{\textbf{Left:} Softmax op weights over all edges in the cell when training the supernet with PPO + "Classic" search space on Dodgeball. Zero op weight is not shown to improve clarity. \textbf{Right:} Sanity check to verify that the supernet eventually achieves a regular training curve, using vanilla IMPALA-CNN as a rough gauge. Both use depths $16 \times 3$. Note that while we show the curve up to 50M steps, we by default discretize $\alpha$ at 25M steps, as op choices have already converged towards a sparser solution. Analogous figure for Rainbow in Appendix \ref{appendix:extended_supernet}.}
\label{fig:ppo_supernet}
\end{figure}

In Figure \ref{fig:ppo_supernet}, we verify that \textbf{training the supernet end-to-end works effectively, even with minimal hyperparameter tuning}. Furthermore, the training only requires at 3x more compute time, with extensive efficiency metrics calculated in Appendix \ref{appendix:efficiency}. However, as mentioned in Subsec. \ref{subsec:sl_vs_rl}, it is unclear whether the right signals are provided to operation routing variables $\alpha$ via the RL training loss, and whether $\alpha$ produces the correct behavior, which we investigate.

Conveniently in Figure \ref{fig:ppo_supernet}, we find that $\alpha$ strongly downweights all base ops (in particular, 5x5 ops) except for the standard Conv3x3+ReLU. This provides an opportunity to understand whether \textbf{$\alpha$ downweights suboptimal ops throughout training.} We confirm this result in Table \ref{table:3x3_vs_5x5} by evaluating standard IMPALA-CNN cells using either purely 3x3 or 5x5 convolutions for the whole network, and demonstrating that the 3x3 setting outperforms the 5x5 setting (especially in limited data, e.g. 200-level training/test regime), suggesting the signaling ability of $\alpha$ on op choice. 

\begin{wraptable}[8]{r}{0.5\textwidth}
\centering
\begin{tabular}{ccc}
\hline
Scenario & Conv 3x3 & Conv 5x5 \\
\hline
Train (Inf. levels) & \textbf{15.1 $\pm$ 2.5} & 13.2 $\pm$ 2.3 \\
\hline
Train (200 levels) & \textbf{12.1 $\pm$ 1.7} & 9.8 $\pm$ 2.1  \\
\hline
Test (from Train) & \textbf{10.2 $\pm$ 2.3} & 5.9 $\pm$ 1.7 \\
\hline
\end{tabular}
\caption{PPO IMPALA-CNN evaluations (mean return at 50M steps) on Dodgeball. Learning curves can be found in Appendix \ref{appendix:softmax_vs_discrete}, Figure \ref{fig:3x3_vs_5x5}.}
\label{table:3x3_vs_5x5}
\end{wraptable}

Answering the converse question is just as important: \textit{Can any supernet train?} The supernet possesses an incredibly dense set of weights, and thus one might wonder whether trainability occurs with any search space or settings. We answer in the negative, where a \textbf{poorly designed supernet can fail.} To show this clearly, we remove all ReLU nonlinearities from the "Classic" search space ops used for PPO, as well as simply freeze $\alpha$ to be uniform for Rainbow, and find both cases produce poor training in Table \ref{table:poor_supernets}. \textbf{Thus, the supernet in RL provides important search signals in terms of reward and $\bm{\alpha}$ during training, especially on the quality of a search space.}

\begin{table}[h]
\centering
\begin{tabular}{c||cc||cc}
\hline
 & \multicolumn{2}{c||}{Rainbow (25M Steps)} & \multicolumn{2}{c}{PPO (50M Steps)} \\
\hline
Scenario & Trainable $\alpha$ & Uniform $\alpha$ & With ReLU & No ReLU \\
\hline
Training (Inf. levels) Reward & \textbf{3.1 $\pm$ 0.5} & 0.9 $\pm$ 0.2 & \textbf{7 $\pm$ 0.9} & 1.9 $\pm$ 0.3 \\
\hline
\end{tabular}
\caption{Supernet training rewards on Dodgeball. Learning curves can be found in Appendix \ref{appendix:supernet_ablation}.}
\label{table:poor_supernets}
\end{table}

\subsection{Discrete Cell Improvement}
\label{subsec:discrete_cell_improvement}

We further must analyze whether discrete cells also improve, as they are used for final evaluation and deployment. Strong supernet performance (via continuous relaxation) does not necessarily imply strong evaluation cell performance (via discretization), due to the existence of \textit{integrality gaps} \citep{rethinking_darts, idarts}. Nevertheless, we demonstrate that even using the default $\delta$ discretization from \citep{darts} leads to both quantitative and \edit{qualitative} improvements.

\begin{wrapfigure}[13]{r}{0.3\textwidth}
\vspace{-6pt}
\includegraphics[width=0.99\linewidth]{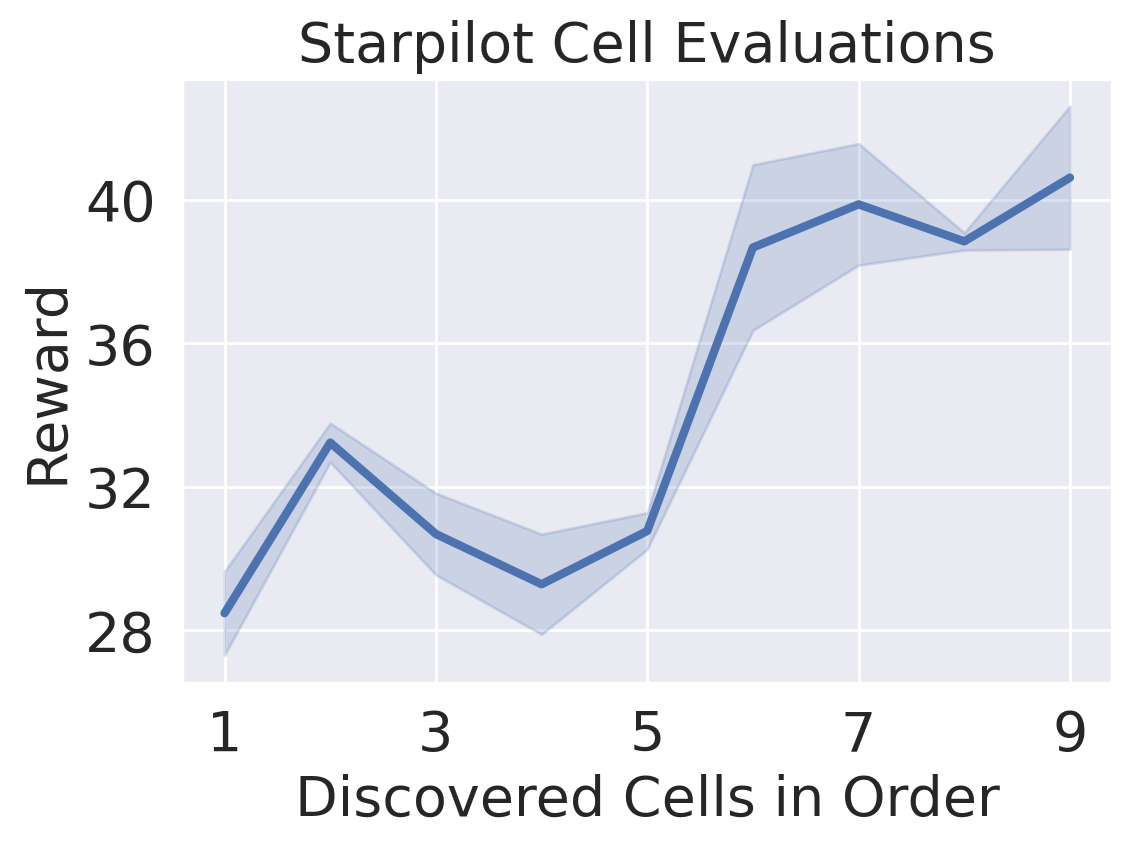}
\caption{Evaluation of 9 distinct discrete cells in order from the trajectory of $\alpha$ on the Starpilot environment when using Rainbow.}
\label{fig:cell_evolve_plot}
\end{wrapfigure}

\paragraph{Quantitative improvement} As $\alpha$ changes during supernet training, so do the outputs of the discretization procedure on $\alpha$. We collect all distinct discrete cells $\{\delta(\alpha_{1}), \delta(\alpha_{2}),\ldots\}$ into a sequence, and evaluate each cell's performance $\max_{\phi} J(\pi_{\phi, \delta(a_{i})}) \> \> \forall i$ via training from scratch for 25M steps (Figure \ref{fig:cell_evolve_plot}). The performance generally improves over time, indicating that supernet optimization selects better cells. However, we find that such behavior can be environment-dependent, as some environments possess less monotonic evaluation curves (see Fig. \ref{fig:dqn-evolve-plots} in Appendix \ref{appendix:discrete_cell}).

\begin{figure}[h]
    \includegraphics[keepaspectratio, width=0.55\textwidth]{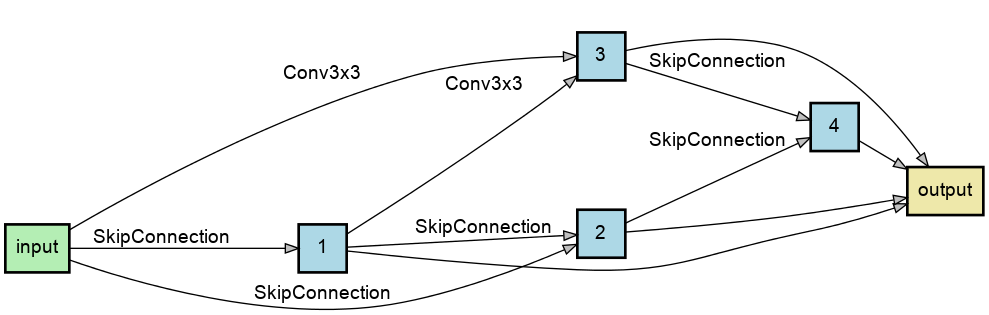}
    \includegraphics[keepaspectratio, width=0.4\textwidth]{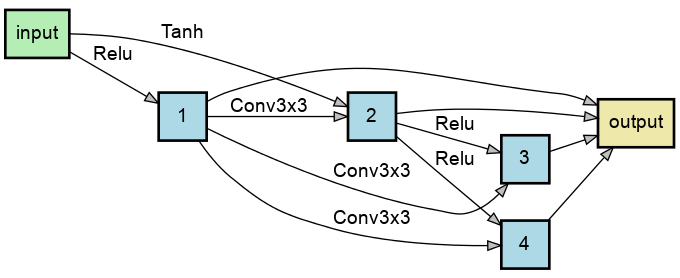}
    \caption{Evolution of discovered cells over a DARTS optimization process. \textbf{Left}: $\delta(\alpha_{start})$ discovered in the early stage which is dominated by skip connections and only linear ops. \textbf{Right}: $\delta(\alpha_{end})$ discovered in the end which possesses several reasonable local structures similar to Conv + ReLU residual connections.}
    \label{fig:cell_evolve}
\end{figure}

\paragraph{Qualitative improvement} We visualize discrete cells $\delta(\alpha_{start}), \delta(\alpha_{end})$ from the start and end of supernet training with Rainbow on the Starpilot environment in Fig. \ref{fig:cell_evolve}. The earlier cell consists of only linear operators is clearly a poor design in comparison to the later cell. We find similar cell evolution results for PPO in Figures \ref{fig:ppo_6_last_node_startpilot_cell_evolve}, \ref{fig:ppo_6_last_node_plunder_cell_evolve} in Appendix \ref{appendix:discrete_cell_evolution}, displaying more sophisticated yet still interpretable changes in cell topology.

\begin{figure}[h]
    \centering
    \setlength\tabcolsep{1.5pt}    
    \begin{tabular}{c c}
    \smallskip
    \begin{subfigure}{0.475\textwidth}
        \centering
        \includegraphics[keepaspectratio, width=0.9\textwidth]{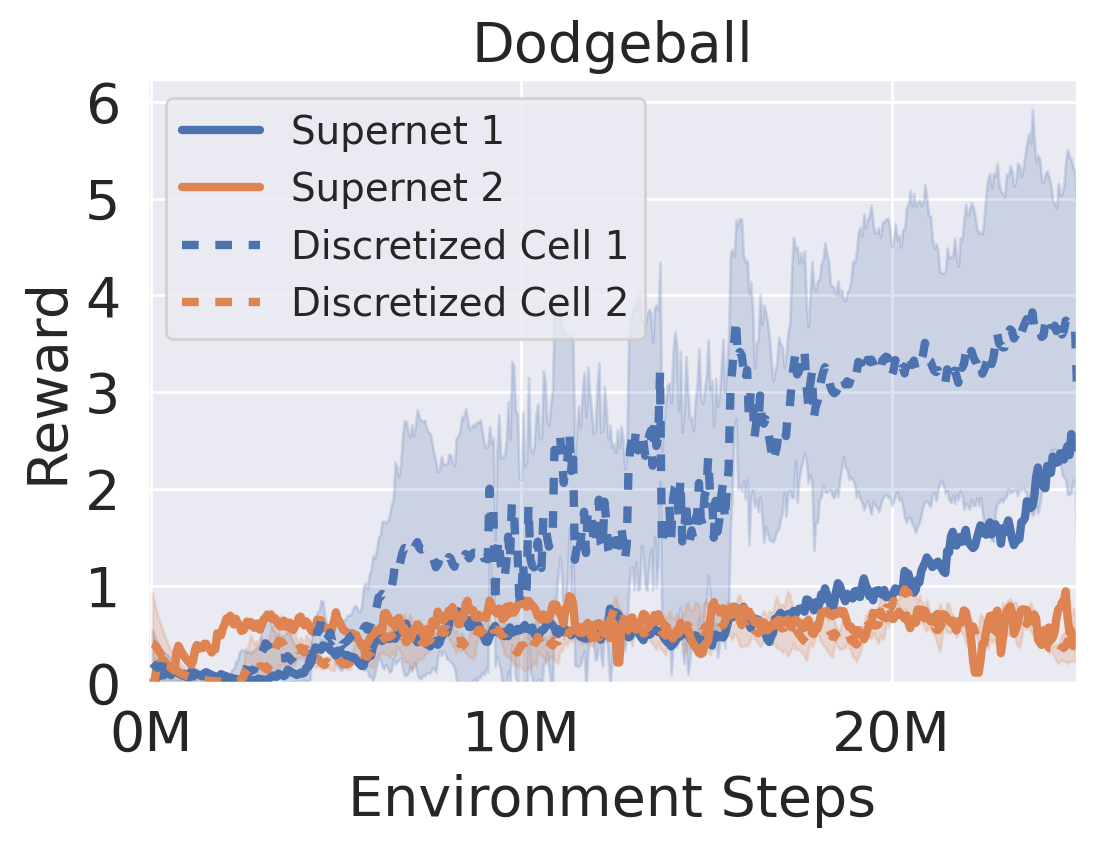}
        \caption{Different supernet runs, with corresponding discretized cell (depths $64 \times 5$) training curves.}
    \end{subfigure}
         &  
    \begin{tabular}{c}
        \smallskip
        \begin{subfigure}[t]{0.475\textwidth}
            \centering
            \includegraphics[keepaspectratio, width=0.9\textwidth]{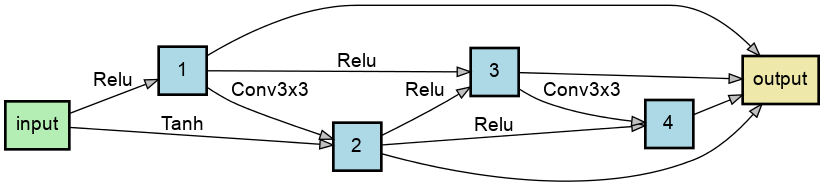}
            \caption{Discretized Cell 1}
        \end{subfigure}
        \\
        \begin{subfigure}[t]{0.475\textwidth}
            \centering
            \includegraphics[keepaspectratio, width=0.9\textwidth]{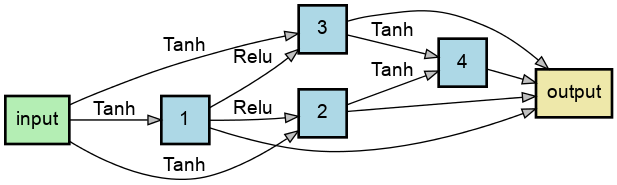}
            \caption{Discretized Cell 2}
        \end{subfigure}        
    \end{tabular}
    \end{tabular}
    \caption{\textbf{(a)} Two different supernets trained on the Dodgeball environment using Rainbow, with corresponding discretized cells evaluated using 3 random seeds. \textbf{(b)} Discretized cell from Supernet 1. Note the similarity to regular Conv3x3 + ReLU designs. \textbf{(c)} Discretized cell from Supernet 2, which uses too many Tanh nonlinearities, known to cause vanishing gradient effects.}
    \label{fig:degenerated-supernet}
\end{figure}

We further answer one of the most common questions in DARTS research: \textit{What is the direct relationship between a supernet and its discrete cell?} In Figure \ref{fig:degenerated-supernet}, we provide two supernet runs along with their corresponding discrete cells, side by side in order to answer this question. While "Supernet 1" is a standard successful training run, "Supernet 2" is a failed run which can commonly occur due to the inherent sensitivity and variance in RL training. As it turns out, this directly leads to differences between their corresponding discrete cells both quantitatively and qualitatively as well, in which "Discretized Cell 1"'s design appears to make sense and train properly, while "Discretized Cell 2" is clearly a suboptimal design, and fails to train at all. Thus, a major reason for why a \textbf{discretized cell may underperform is if its corresponding supernet fails to learn}. 

We investigate further in Appendix \ref{subsec:correlation}, Figure \ref{fig:supernet_vs_discrete_rainbow}, and find that supernet and discrete cell rewards are indeed correlated, after adjusting for environment-dependent factors, suggesting that search quality can be improved via both better supernet training \citep{autodrac, model_rl_hp_tuning}, as well as better discretization procedures \citep{darts_plus, rethinking_darts}.

\section{Conclusions, Limitations, and Broader Impact Statement}
\label{sec:conclusion}
\paragraph{Conclusion} Even though RL uses complex loss functions defined over nonstationary data, we empirically showed that nonetheless, DARTS is capable of improving policy architectures in a minimally invasive and efficient way (only 3x more compute time) across several different algorithms (PPO, Rainbow, SAC) and environments (Procgen, DM-Control). Our paper is the first to have comprehensively provided evidence for the applicability of softmax/gradient-based architecture search outside of standard classification and SL.

\paragraph{Limitations} To avoid confounding factors and for simplicity, our paper uses the default DARTS method. However, we outline multiple possible improvements in Appendix \ref{appendix:possible_improvements}, as RL is a completely new frontier for which to understand softmax routing and continuous relaxation techniques. 

\paragraph{Future Work and Broader Impact} In this paper, we have provided concrete evidence that architecture search can be conducted practically and efficiently in RL, via DARTS. We believe that this could start important initiatives into finding better and more efficient \citep{proxyless_nas} architectures for large-scale robotics \citep{sim2realviasim2sim, visionary}, transferable architectures in offline RL \citep{offline_rl_tutorial}, as well as RNNs for memory \citep{neural_episodic_control, rl_working_episodic_memory, rnn_atari} and adaptation \citep{rl_squared, l2rl}. \edit{Other NAS methods' applicability in RL may also be investigated, especially ones which utilize blackbox optimization controllers, such as multi-trial evolution \citep{reg_evo} or ENAS \citep{enas}.}

\edit{Furthermore, we believe our work's potential negative impacts are generally equivalent to ones found in general NAS methods, such as sacrificing model interpretability in order to achieve higher objectives. For the field of RL specifically, this may warrant more attention in AI safety when used for real world robotic pipelines. Furthermore, as with any NAS research, the initial phase of discovery and experimentation may contribute to carbon emissions due to the computational costs of extensive tuning. However, this is usually a means to an end, such as an efficient search algorithm, which this paper proposes with no extra hardware costs.}


\medskip
\small

\clearpage

\section{Reproducibility Checklist}

\begin{enumerate}
\item For all authors\dots
  \begin{enumerate}
  \item Do the main claims made in the abstract and introduction accurately
    reflect the paper's contributions and scope?
    \answerYes{We have comprehensively demonstrated the behavior of all components in DARTS when applied inside an RL pipeline.}
  \item Did you describe the limitations of your work?
    \answerYes{Yes, see Section \ref{sec:conclusion}.}
  \item Did you discuss any potential negative societal impacts of your work?
    \answerYes{Yes, see Section \ref{sec:conclusion}.}
  \item Have you read the ethics review guidelines and ensured that your paper
    conforms to them?
    \answerYes{Yes, see Section \ref{sec:conclusion}.}
  \end{enumerate}
\item If you are including theoretical results\dots
  \begin{enumerate}
  \item Did you state the full set of assumptions of all theoretical results?
    \answerNA{No theoretical results provided.}
  \item Did you include complete proofs of all theoretical results?
    \answerNA{No proofs provided.}
  \end{enumerate}
\item If you ran experiments\dots
  \begin{enumerate}
  \item Did you include the code, data, and instructions needed to reproduce the
    main experimental results, including all requirements (e.g.,
    \texttt{requirements.txt} with explicit version), an instructive
    \texttt{README} with installation, and execution commands (either in the
    supplemental material or as a \textsc{url})?
    \answerYes{We have included the core code for creating a DARTS supernet and discrete cell, as well as training code for Procgen on both PPO and Rainbow, \edit{and addition added the modified files from an open-source variant of SAC on DM-Control.} We have also added the README on guidance for code organization and running experiments, as well as a \texttt{requirements.txt}. The code can be found at \url{https://github.com/google/brain_autorl/tree/main/rl_darts}.}
  \item Did you include the raw results of running the given instructions on the
    given code and data?
    \answerYes{We have provided the data related to the core results of this paper.}
  \item Did you include scripts and commands that can be used to generate the
    figures and tables in your paper based on the raw results of the code, data,
    and instructions given?
    \answerYes{We have provided relevant figure-plotting utilities in the code submission.}
  \item Did you ensure sufficient code quality such that your code can be safely
    executed and the code is properly documented?
    \answerYes{The code was formatted by a strict automatic Python lint-checker, as well as reviewed by other individuals. The code also provides tests for modules, which can be used to understand the intended use.}
  \item Did you specify all the training details (e.g., data splits,
    pre-processing, search spaces, fixed hyperparameter settings, and how they
    were chosen)?
    \answerYes{We discuss the explicit hyperparameters in Appendix \ref{appendix:hyperparameters}, as well as search space details in Section \ref{sec:experiments}.}
  \item Did you ensure that you compared different methods (including your own)
    exactly on the same benchmarks, including the same datasets, search space,
    code for training and hyperparameters for that code?
    \answerYes{We ensured the same evaluation protocol via making sure all depths remain the same ($16$ for supernets, $64$ for discrete cells, and $D=3$ layers for cheaper large-scale runs or $D=5$ for fine-grained A/B testing). We also used the exact same hyperparameters for all discrete/baseline runs, while only needing to change minibatch size + learning rate in accomodate networks with larger GPU memory sizes.}
  \item Did you run ablation studies to assess the impact of different
    components of your approach?
    \answerYes{For ablation studies, we provided a large set of studies as seen throughout Subsections \ref{subsec:supernet_training} and \ref{subsec:discrete_cell_improvement} as well as in the Appendix.}
  \item Did you use the same evaluation protocol for the methods being compared?
    \answerYes{In terms of comparing NAS methods, we indeed controlled for confounding factors by using the same hardware, as shown in Table \ref{table:efficiency} in Appendix \ref{appendix:efficiency}. We further performed side-by-side comparisons for RL training curves using environment steps, which is standard.}
  \item Did you compare performance over time?
    \answerYes{Table \ref{table:efficiency} in Appendix \ref{appendix:efficiency} contains wall-clock time comparisons. In terms of performance comparisons over time, we showed the performance of RL-DARTS's discrete cells over the search procedure / supernet training in Figure \ref{fig:cell_evolve_plot}.}
  \item Did you perform multiple runs of your experiments and report random seeds?
    \answerYes{For seeded runs, as standard in RL, we performed 3-seeded runs for all experiments.}
  \item Did you report error bars (e.g., with respect to the random seed after
    running experiments multiple times)?
    \answerYes{For seeded runs, as standard in RL, we performed 3-seeded runs for all experiments and reported error bars.}
  \item Did you use tabular or surrogate benchmarks for in-depth evaluations?
    \answerYes{We extensively covered the comparisons against random search in Subsection \ref{subsec:end_to_end}, over all 16 games as well as a large scale 100 random cell comparison in Figure \ref{fig:games-random-cell-hist}. }
  \item Did you include the total amount of compute and the type of resources
    used (e.g., type of \textsc{gpu}s, internal cluster, or cloud provider)?
    \answerYes{See Table \ref{table:efficiency} in Appendix \ref{appendix:efficiency} for GPU and compute time used for all evaluations. We applied this to all 16 games on Procgen.}
  \item Did you report how you tuned hyperparameters, and what time and
    resources this required (if they were not automatically tuned by your AutoML
    method, e.g. in a \textsc{nas} approach; and also hyperparameters of your
    own method)?
    \answerYes{This is written in detail in Appendix \ref{appendix:hyperparameters} as well as Section \ref{sec:experiments}.}
  \end{enumerate}
\item If you are using existing assets (e.g., code, data, models) or
  curating/releasing new assets\dots
  \begin{enumerate}
  \item If your work uses existing assets, did you cite the creators?
    \answerYes{We have cited Procgen and DM-Control.}
  \item Did you mention the license of the assets?
    \answerNA{Both are publicly licensed and freely available.}
  \item Did you include any new assets either in the supplemental material or as
    a \textsc{url}?
    \answerYes{We included the relevant publicly available code for environments and algorithms in Section \ref{appendix:hyperparameters}.}
  \item Did you discuss whether and how consent was obtained from people whose
    data you're using/curating?
    \answerNA{Benchmarks are publicly available, consent not needed.}
  \item Did you discuss whether the data you are using/curating contains
    personally identifiable information or offensive content?
    \answerNA{No personally identifiable information or offensive content.}
  \end{enumerate}
\item If you used crowdsourcing or conducted research with human subjects\dots
  \begin{enumerate}
  \item Did you include the full text of instructions given to participants and
    screenshots, if applicable?
    \answerNA{No research was conducted on human subjects or crowdsourcing.}
  \item Did you describe any potential participant risks, with links to
    Institutional Review Board (\textsc{irb}) approvals, if applicable?
    \answerNA{No research was conducted on human subjects or crowdsourcing.}
  \item Did you include the estimated hourly wage paid to participants and the
    total amount spent on participant compensation?
    \answerNA{No research was conducted on human subjects or crowdsourcing.}
  \end{enumerate}
\end{enumerate}

\clearpage
\bibliography{references}
\bibliographystyle{apalike}

\clearpage
\appendix

\input{appendix}

\end{document}

%% file: math_commands.tex
\def\eqref#1{equation~\ref{#1}}

\def\1{\bm{1}}

\DeclareMathAlphabet{\mathsfit}{\encodingdefault}{\sfdefault}{m}{sl}
\SetMathAlphabet{\mathsfit}{bold}{\encodingdefault}{\sfdefault}{bx}{n}

\DeclareMathOperator*{\argmax}{arg\,max}
\DeclareMathOperator*{\argmin}{arg\,min}

%% file: appendix.tex
\LARGE
\textsc{Appendix}
\normalsize

\section{Efficiency Metrics}
\label{appendix:efficiency}
We provide computational efficiency metrics in Table \ref{table:efficiency}, where we find that the practical wall-clock time required for training the supernet (i.e. the search cost) is very comparable with DARTS in SL \citep{darts}, requiring only a few GPU days. We do note that unlike SL where the vast majority of the cost is due to the network, RL time cost is partially based on non-network factors such as environment simulation, and thus wall-clock times may change depending on specific implementation.

\begin{table}[h]
\centering
\begin{tabular}{l*{4}{c}r}
    \toprule
    \textbf{Network} 
    & \textbf{Training Cost in GPU Days (w/ specific algorithm)} \\
    \midrule
    IMPALA-CNN 
    & 1 (PPO), 0.5 (Rainbow) \\
    "Classic" Supernet  
    & 2.5 (PPO) \\
    "Micro" Supernet 
    & 1.5 (Rainbow) \\
    CIFAR-10 Supernet \citep{darts} 
    & 4 (SL/Original DARTS) \\
    \bottomrule
    \\
\end{tabular}
\vspace{-10pt}
\caption{Computational efficiency in terms of wall-clock time, achieved on a V100 GPU. For the RL cases (PPO + Rainbow), all networks use depths of $16 \times 3$. Training cost in RL is defined as the wallclock time taken to reach 25M steps, rounded to the nearest 0.5 GPU day. We have also included reported time for DARTS in SL \citep{darts} as comparison.}
\label{table:efficiency}
\end{table}

\section{Extended Supernet Training Results}
\label{appendix:extended_supernet}
Following Subsection \ref{subsec:supernet_training}, we also present a similar figure for Rainbow below, for completeness.

\begin{figure}[h]
    \includegraphics[keepaspectratio, width=0.8\textwidth]{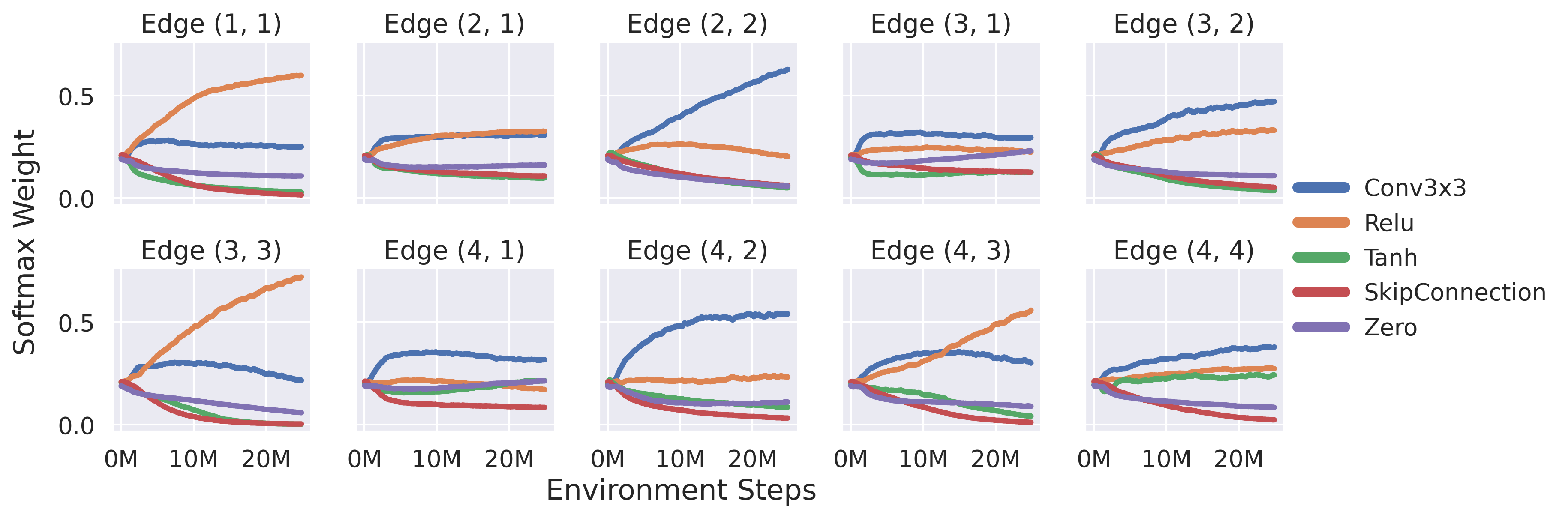}
    \includegraphics[keepaspectratio, width=0.19\textwidth]{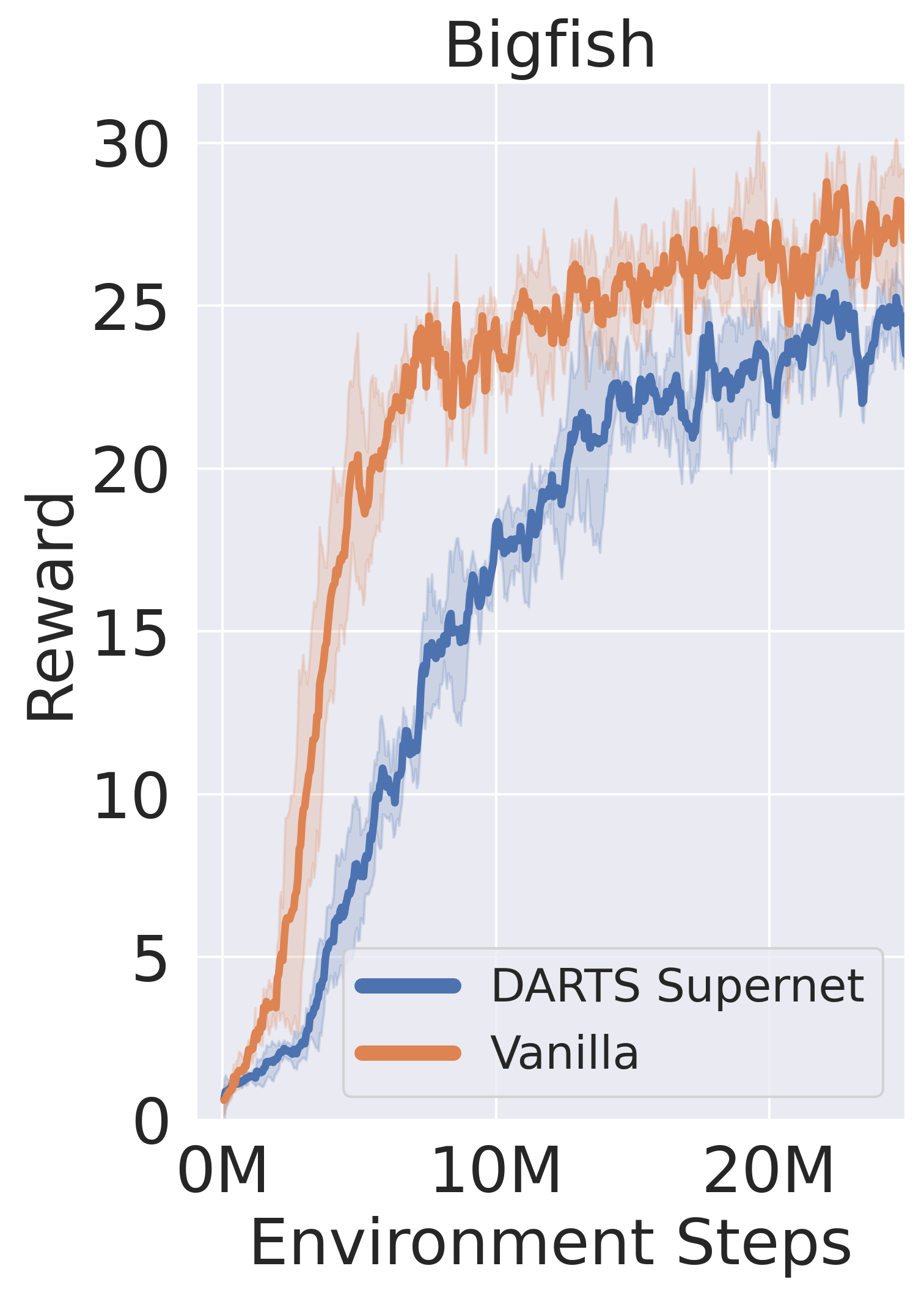}
    \caption{Analogous settings with Figure \ref{fig:ppo_supernet} using Rainbow + "Micro" search space. \textbf{Left:} Softmax weights when training Rainbow with infinite levels on Bigfish, also converging towards a sparser solution. \textbf{Right:} Sanity check for supernet when using Rainbow.}
    \label{fig:rainbow_supernet}
\end{figure}

\section{What Affects Supernet Training?}
\label{appendix:supernet_ablation}
Given the positive training results we demonstrated in the main body of the paper, one may wonder, \textit{can any supernet, no matter how poorly designed or setup, still train well in the RL setting?} If so, this would imply that the search method would not be producing meaningful, but instead, random signals.

We refute this hypothesis by performing ablations over our supernet training in order to have a better understanding of what components affect its performance. We ultimately show that the search space and architecture variables play a very significant role in its optimization, thus validating our method.

\subsection{Role of Search Space}
\begin{figure}[h]
    \center
    \includegraphics[keepaspectratio, width=0.6\textwidth]{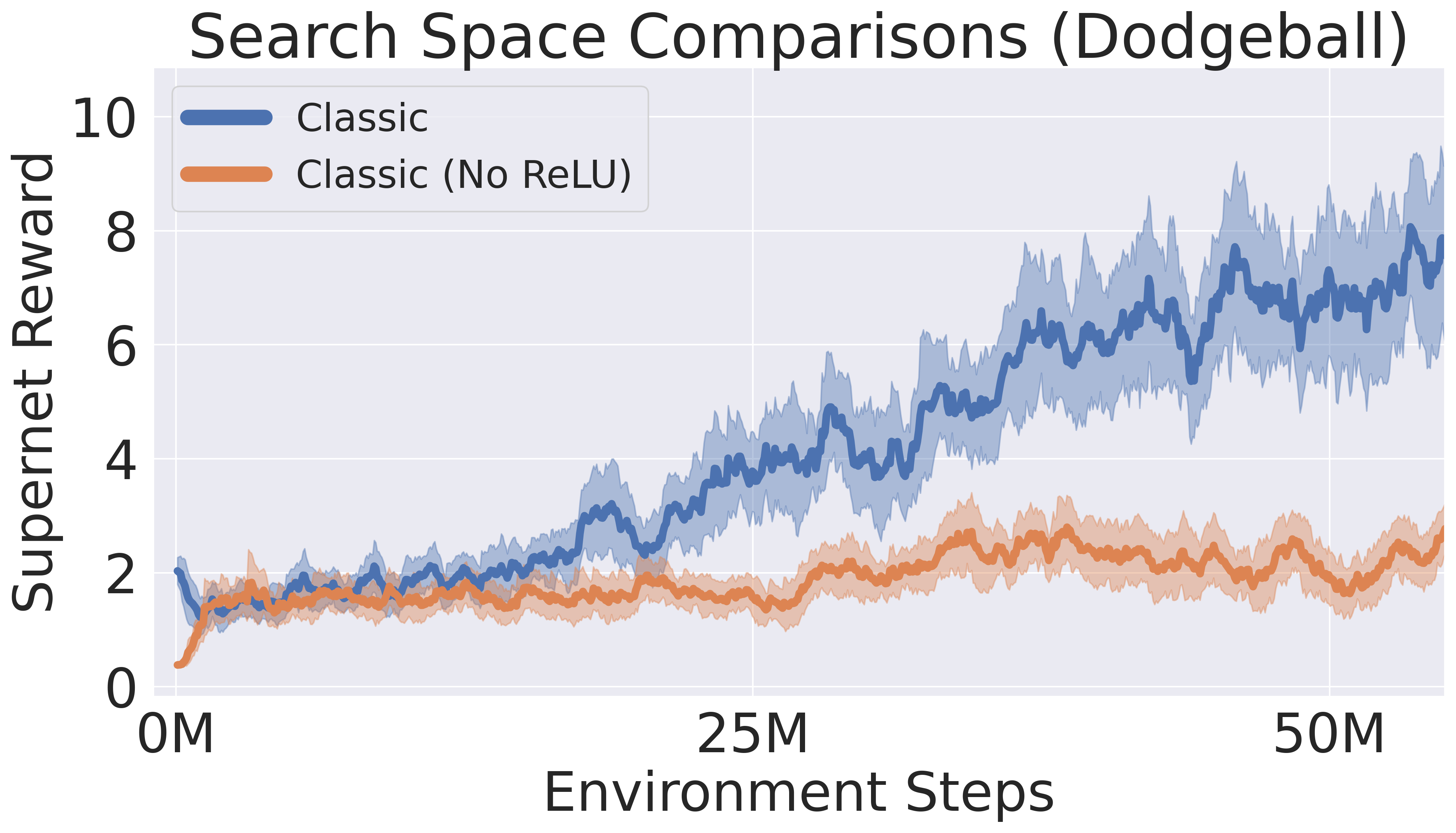}
\caption{Supernet training using PPO on Dodgeball with infinite levels, when using the "Classic" search space with/without ReLU nonlinearities, under the same hyperparameters.}
\label{fig:ppo_search_space_comparisons}
\end{figure}

We remove the ReLU nonlinearities from the "Classic" search space, so that $\mathcal{O}_{base}$ = $\{$Conv3x3, Conv5x5, Dilated3x3, Dilated5x5$\}$ and thus the DARTS cell consists of only linear operations. As shown in Fig. \ref{fig:ppo_search_space_comparisons}, this leads to a dramatic decrease in supernet performance, providing evidence that the search space matters greatly. 

\subsection{Uniform Architecture Variables}
\begin{figure}[h]
    \center
    \includegraphics[keepaspectratio, width=0.6\textwidth]{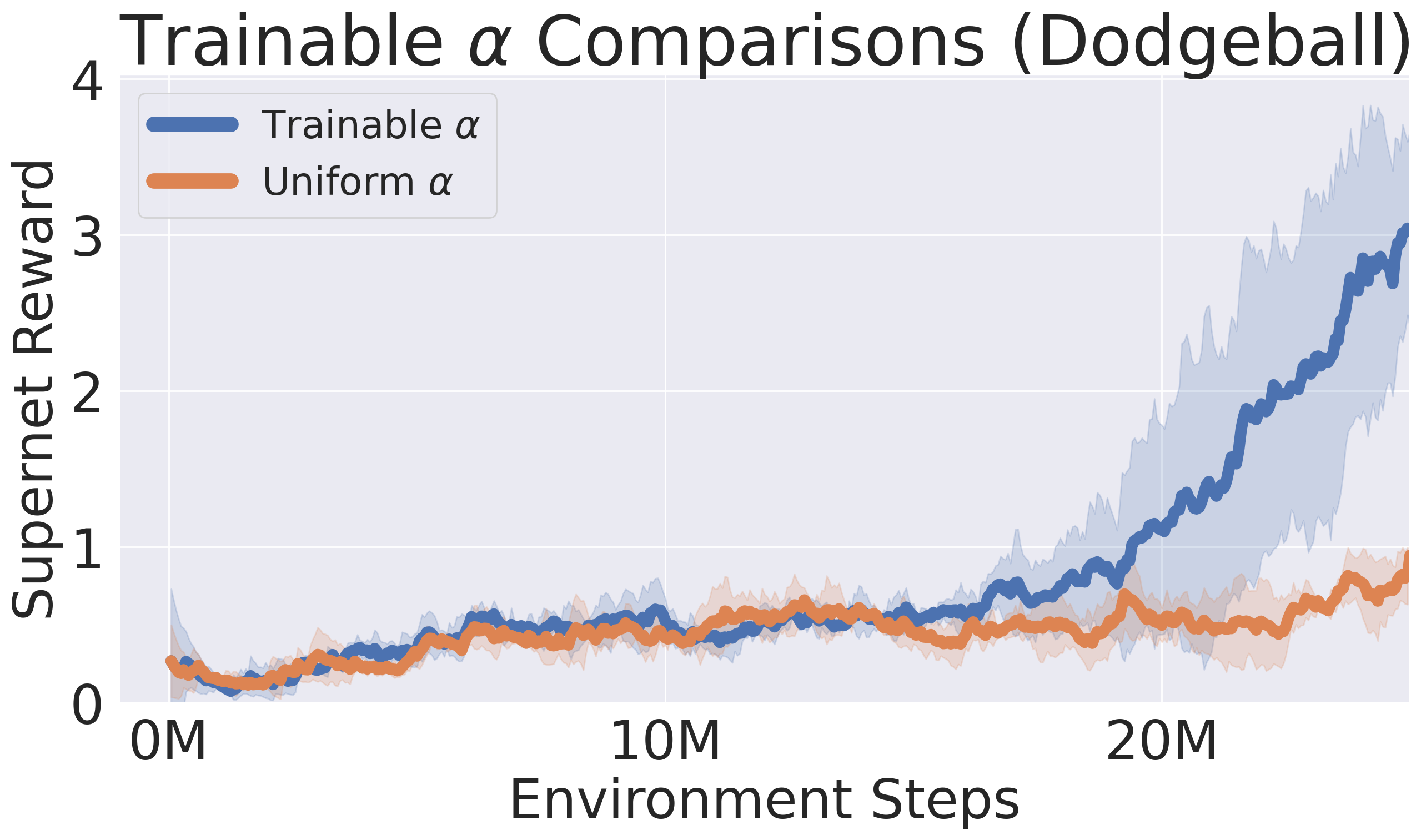}
\caption{Supernet training using Rainbow on Dodgeball with infinite levels, when using the "Micro" search space with/without trainable architecture variables $\alpha$, under the same hyperparameters.}
\label{fig:dqn_uniform_arch_var}
\end{figure}

We further demonstrate the importance of the architecture variables $\alpha$ on training. We see that in Fig. \ref{fig:dqn_uniform_arch_var}, freezing $\alpha$ to be uniform throughout training makes the Rainbow agent unable to train at all. This suggests that it is crucial for $\alpha$ to properly route useful operations throughout the network.

\clearpage

\section{What Affects Discrete Cell Performance?}
\label{appendix:discrete_cell}

\subsection{Softmax Weights vs Discretization}
\label{appendix:softmax_vs_discrete}
As seen from Figure \ref{fig:ppo_supernet} in the main body of the paper, the DARTS supernet strongly downweights Conv5x5+ReLU operations when using the "Classic" search space with PPO. In order to verify the predictive power of the softmax weights, as a proxy, we thus also performed evaluations when using purely 3x3 or 5x5 convolutions on a large IMPALA-CNN with $64 \times 5$ depths. We see that the Conv3x3 setting indeed outperforms Conv5x5, corroborating the results in which during training, $\alpha$ strongly upweights the Conv3x3+ReLU operation and downweights Conv5x5+ReLU.

\begin{figure}[h]
    \center
    \includegraphics[keepaspectratio, width=0.7\textwidth]{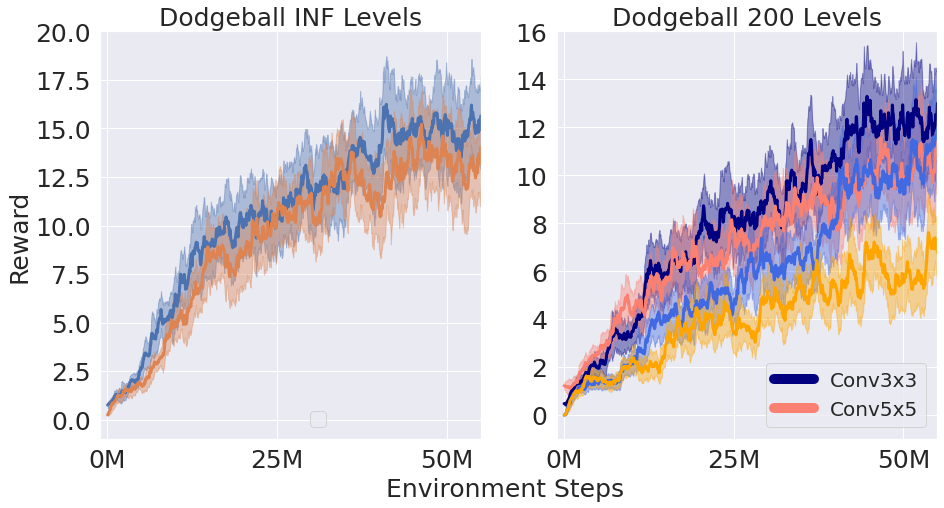}
\caption{Large IMPALA-CNNs evaluated on Dodgeball using either Infinite or 200 levels with PPO. For the 200 level setting, lighter colors correspond to test performance.}
\label{fig:3x3_vs_5x5}
\end{figure}


\subsection{Discrete Cell Evolutions} 
\label{appendix:discrete_cell_evolution}
Along with Figure \ref{fig:cell_evolve} in the main body of the paper, we also compare extra examples of discretizations before and after supernet training, to display reasonable behaviors induced by the trajectory of $\alpha$.

\begin{figure}[h]
    \includegraphics[keepaspectratio, width=0.47\textwidth]{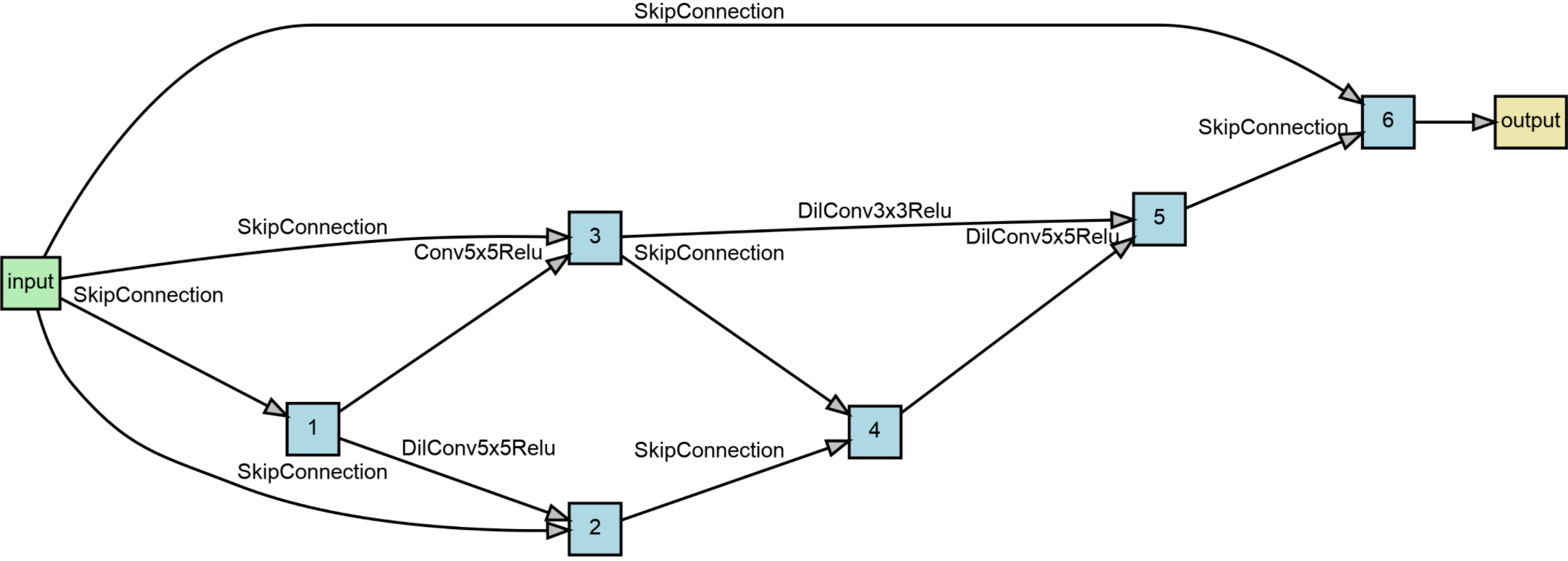}
    \includegraphics[keepaspectratio, width=0.53\textwidth]{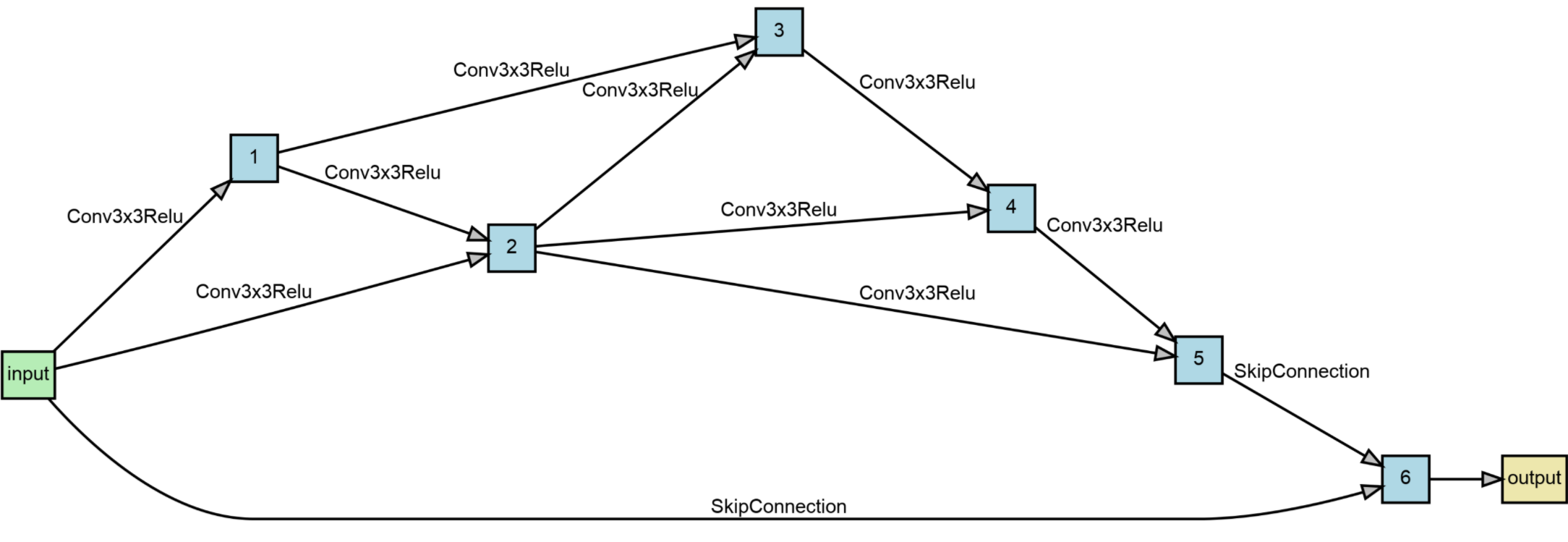}
    \caption{Comparison of discretized cells before and after supernet training, on Starpilot using PPO with $I=6$ nodes.}
    \label{fig:ppo_6_last_node_startpilot_cell_evolve}
\end{figure}

\begin{figure}[h]
    \includegraphics[keepaspectratio, width=0.43\textwidth]{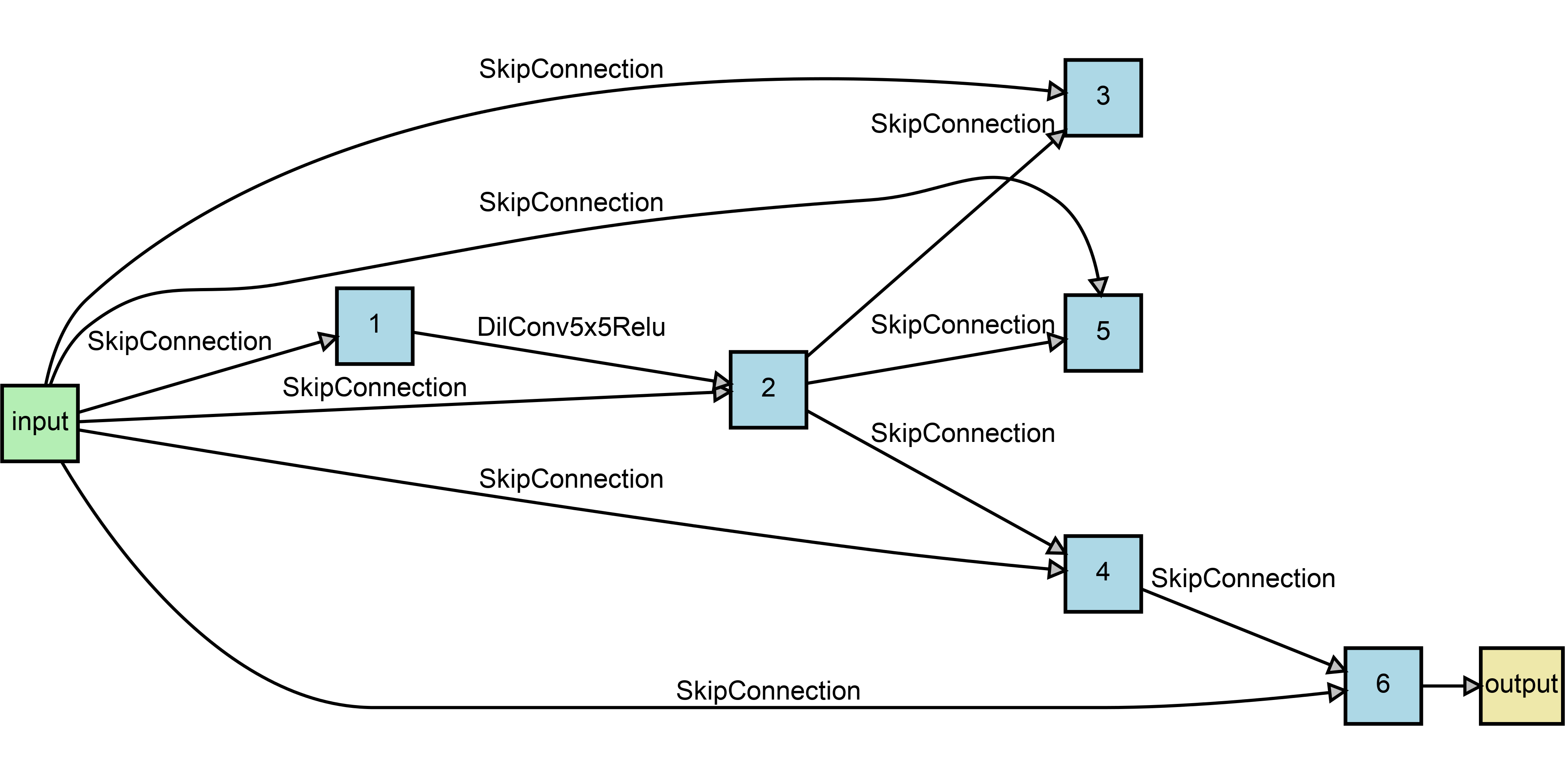}
    \includegraphics[keepaspectratio, width=0.57\textwidth]{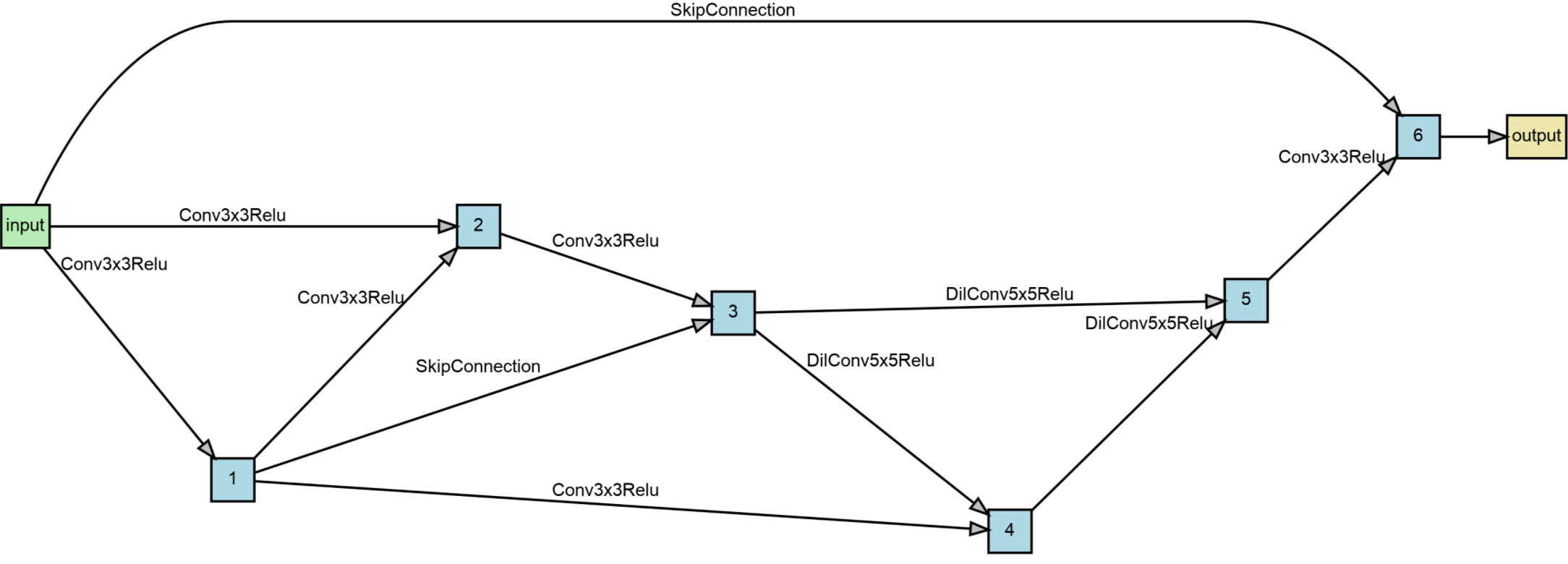}
    \caption{Comparison of discretized cells before and after supernet training, on Plunder using PPO with $I=6$ nodes.}
    \label{fig:ppo_6_last_node_plunder_cell_evolve}
\end{figure}
In Figure \ref{fig:ppo_6_last_node_startpilot_cell_evolve}, we use PPO with the "Classic" search space, but instead use $(N, R, I) = (1,0,6)$ along with outputting the last node (instead of concatenation with a Conv1x1 for the output) to allow a larger normal cell search space and graph topology. In Figure \ref{fig:ppo_6_last_node_plunder_cell_evolve}, the discretized cell initially uses a large number of skip connections as well as dead-end nodes. However, at convergence, it eventually utilizes all nodes to compute the final output. Curiously, we find that the skip connection between the input and output appears commonly throughout many searches.

\begin{figure}[h]
    \includegraphics[keepaspectratio, width=0.95\textwidth]{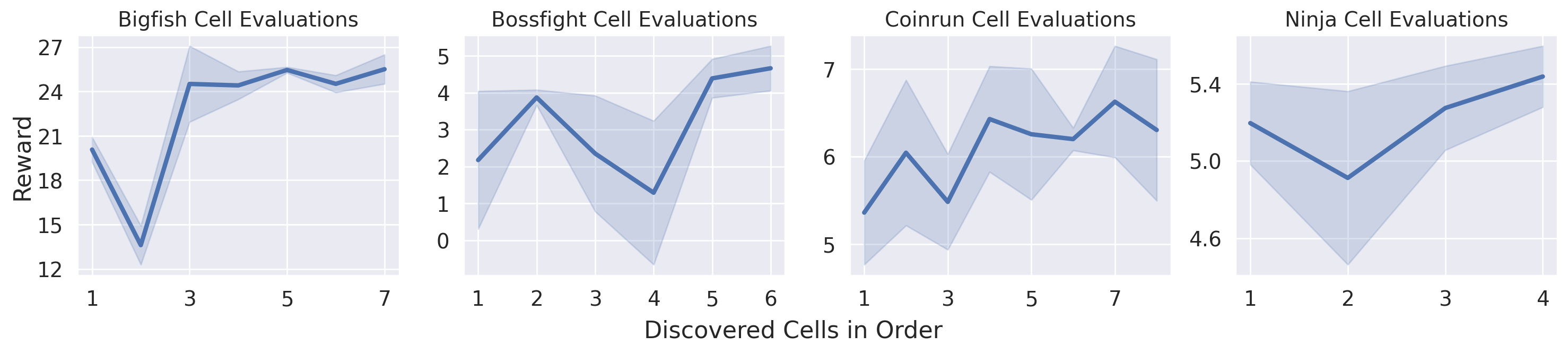}
    \caption{Evaluated discretized cells discovered throughout training the supernet with Rainbow. To save computation, we evaluate every 2nd cell that was discovered.}
    \label{fig:dqn-evolve-plots}
\end{figure}

For the Rainbow setting, in Figure \ref{fig:cell_evolve_plot} in the main body of the paper, we saw that when the search process is successful, the supernet's training trajectory induces discretized cells which improve evaluation performance as well. The cells discovered later generally perform better than cells discovered earlier in the supernet training process. In Figure \ref{fig:dqn-evolve-plots}, we show more examples of such evaluation curves.

\subsection{Correlation Between Supernet and Discretized Cells}
\label{subsec:correlation}
\begin{figure}[h]
    \centering
    \includegraphics[keepaspectratio, width=0.95\textwidth]{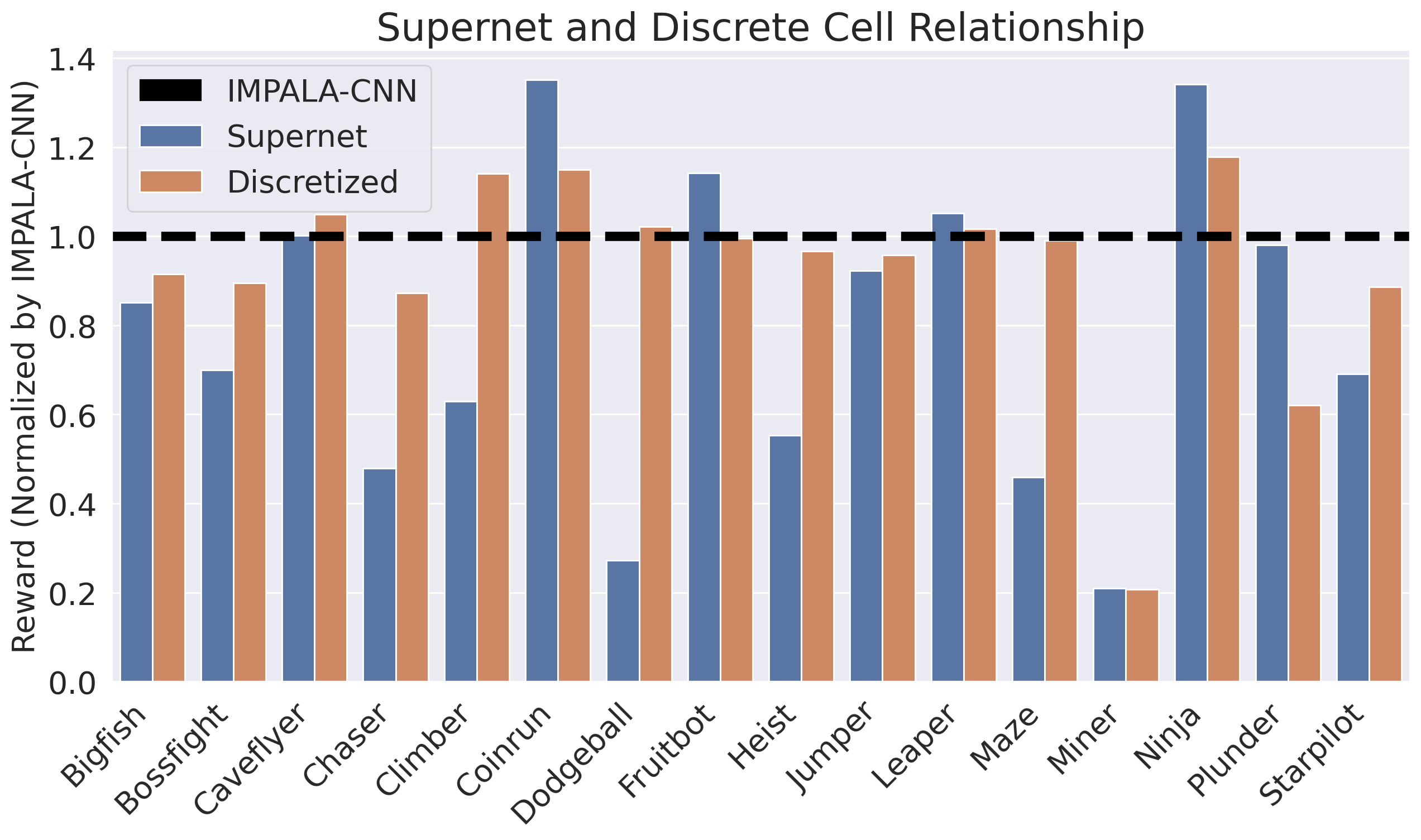}
    \caption{Supernet and their corresponding discrete cell rewards across all environments in Procgen using Rainbow, after normalizing using IMPALA-CNN's performances. Thus, the black dashed line at 1.0 corresponds to IMPALA-CNN.}
    \label{fig:supernet_vs_discrete_rainbow}
\end{figure}

Given that the discretized cell only explicitly depends on architecture variables $\alpha$ and not necessarily model weights $\theta$, one may wonder: \textit{Is there a relationship between the rewards of the supernet and of its corresponding discretized cell?} For instance, the degenerate/underperformance setting mentioned in Section \ref{subsec:discrete_cell_improvement} and Appendix \ref{appendix:discrete_cell} can be thought of as an extreme scenario. At the same time, there could be an integrality gap, where there the discretization process $\delta(\alpha)$ produces cells which give different rewards than the supernet.

In order to make such an analysis comparing rewards, we first must prevent confounding factors arising from Rainbow's natural performance on an environment regardless of architecture. We thus first divide the supernet and discrete cell scores by the score obtained by the IMPALA-CNN baseline, where the baseline and discrete cells all used depths of $64 \times 3$.

In Figure \ref{fig:supernet_vs_discrete_rainbow}, when using Rainbow and observing across environments, we find both high correlation and also integrality gaps: for some environments such as Ninja and Coinrun, there is a significant correlation between supernet and discrete cell rewards, while for other environments such as Dodgeball, there is a significant gap. This suggests that search quality can be improved via both better supernet training such as using hyperparameter tuning or data augmentation \citep{autodrac}, as well as better discretization procedures such as early stopping and stronger pruning \citep{darts_minus, darts_plus, rethinking_darts}.

\section{Numerical Scores}
\label{sec:numerical}
In Tables \ref{ppo_clasic_eval} and \ref{rainbow_micro_eval}, we display the average normalized reward after 25M steps, as standard in Procgen \citep{procgen}, for a subset of environments in which RL-DARTS performs competitively. The normalized reward for each environment is computed as $R_{norm} = (R - R_{min})/(R_{max} - R_{min})$ where $R_{max}$ and $R_{min}$ are calculated using a combination of theoretical maximums and PPO-trained agents, and can be found in \citep{procgen}.


\begin{table}[h]
    \centering
        \smallskip

        \subfloat[PPO + Classic]{
        \begin{tabular}{l*{5}{c}r}
            \toprule
            \textbf{Env} & \textbf{IMPALA-CNN Baseline} & \textbf{RL-DARTS (Discrete)} & \textbf{Random Search} \\
            \midrule
            Bigfish & \textbf{0.60} & \textbf{0.60} & 0.42 \\
            Bossfight & 0.75 & 0.73 & \textbf{0.81} \\
            Caveflyer & 0.75 & 0.47 & \textbf{0.85} \\
            Chaser & \textbf{0.71} & 0.55 & 0.15 \\
            Climber & 0.69 & \textbf{0.90} & 0.61 \\
            Coinrun & \textbf{0.91} & 0.53 & 0.8 \\
            Dodgeball & 0.53 & \textbf{0.59} & 0.29 \\
            Fruitbot & \textbf{0.92} & \textbf{0.93} & 0.83 \\
            Heist & 0.72 & \textbf{0.89} & 0.38 \\
            Jumper & 0.62 & 0.76 & \textbf{1.0} \\
            Leaper & 0.2 & \textbf{0.28} & -0.28 \\
            Maze & \textbf{1.0} & \textbf{1.0} & 0.0 \\
            Miner & 0.74 & \textbf{0.85} & 0.70 \\
            Ninja & \textbf{0.87} & 0.69 & 0.38 \\
            Plunder & 0.57 & \textbf{0.76} & 0.43 \\
            Starpilot & 0.71 & \textbf{0.73} & 0.40 \\
            \bottomrule
            \\
        \end{tabular}
        \label{ppo_classic_eval}
        }

        \hfill
        \smallskip

        \subfloat[Rainbow + Micro]{
        \begin{tabular}{l*{5}{c}r}
            \toprule
            \textbf{Env} & \textbf{IMPALA-CNN Baseline} & \textbf{RL-DARTS (Discrete)} & \textbf{Random Search} \\
            \midrule
            Bigfish & \textbf{0.71} & 0.65 & 0.60 \\
            Bossfight & \textbf{0.54} & \textbf{0.48} & 0.45 \\
            Caveflyer & -0.05 & \textbf{-0.03} & \textbf{-0.01} \\
            Chaser & \textbf{0.30} & \textbf{0.26} & 0.22 \\
            Climber & \textbf{-0.04} & \textbf{-0.02} & \textbf{-0.05} \\
            Coinrun & 0.06 & \textbf{0.21} & 0.15 \\
            Dodgeball & \textbf{0.57} & \textbf{0.59} & -0.05 \\
            Fruitbot & \textbf{0.68} & \textbf{0.68} & \textbf{0.70} \\
            Heist & \textbf{-0.48} & \textbf{-0.49} & \textbf{-0.47} \\
            Jumper & \textbf{0.21} & \textbf{0.18} & 0.17 \\
            Leaper & -0.07 & -0.07 & \textbf{-0.03} \\
            Maze & \textbf{0.76} & \textbf{0.74} & 0.51 \\
            Miner & \textbf{0.35} & -0.03 & 0.11 \\
            Ninja & 0.03 & 0.13 & \textbf{0.28} \\
            Plunder & \textbf{0.14} & 0.02  & 0.03 \\
            Starpilot & \textbf{0.91} & 0.80 & 0.76 \\
            \bottomrule
            \\
        \end{tabular}
        \label{rainbow_micro_eval}
        }

    \caption{Normalized Rewards in ProcGen across different search methods, evaluated at 25M steps with depths $64 \times 3$. Largest scores on the specific environment (as well as values within 0.03 of the largest) are \textbf{bolded}.}
\end{table}

\clearpage

\section{Possible Improvements for Future Work}
\label{appendix:possible_improvements}
These include:

\begin{enumerate}
\item Discretization Changes: One may consider discretization based on the total reward $J(\pi_{\theta, \alpha})$, which may provide a better signal for the correct discrete architecture. This is due to the fact that the relative strengths of operation weights from $\alpha$ may not correspond to the best choices during discretization. \citep{rethinking_darts} considers iteratively pruning edges from the supernet based on maximizing validation accuracy changes. For RL, this would imply a variant of discretization dependent on multiple calculations of $J(\pi_{\theta^{*}, \delta_{1}(\alpha^{*})}) - J(\pi_{\theta^{*}, \delta_{2}(\alpha^{*})})$ where $\theta^{*}$ consists the weights obtained during supernet training, as well as fine-tuning $J(\pi_{\theta^{*}, \delta_{1}(\alpha^{*})})$ at every pruning step. These changes, in addition to the inherently noisy evaluations of $J(\cdot)$, greatly increase the complexity of the discretization procedure, but are worth exploring in future work.

\item Changing the Loss / Regularization: Throughout this paper, we have found that vanilla DARTS is able to train by simply optimizing $\alpha$ with respect to the loss, even though in principle, the loss is not strongly correlated to the actual reward in RL. Thus, it is curious to understand whether loss-based metrics or modifications may help improve RL-DARTS. One such modification is based on the observation that certain RL losses may not be required for training $\alpha$. In PPO, the entropy loss of $\pi_{\theta}$ may not be necessary or useful for improving the search quality of $\alpha$, and thus it may be better to perform a two step update by providing a different loss for $\alpha$. One may also consider searching for two separate encoders via two supernets, since both PPO and Rainbow feature separate networks, e.g. the policy $\pi_{\theta_{1}, \alpha_{1}}$ and value function $V_{\theta_{2}, \alpha_{2}}$ for PPO and advantage function $A_{\theta_{1}, \alpha_{1}}$ and value function $V_{\theta_{2}, \alpha_{2}}$ for Rainbow.

\item Signaling Metrics and Early Stopping: Observing metrics throughout training allows for early stopping, which can reduce search cost and provide better discrete cells. This includes metrics such as the strength of certain operation weights \citep{darts_plus} as well as the Hessian with respect to $\alpha$ throughout training, i.e. $\nabla_{\alpha}^{2} \mathcal{L}(\theta, \alpha)$ as found in \citep{understanding_robustifying_darts}. Furthermore, inspired by performance prediction methods \citep{nas_without_training, neural_architecture_optimization}, one may analyze metrics such as the Jacobian Covariance, via the score defined to be the $-\sum_{i=1}^{B} \left[\log(\sigma_{i} + \varepsilon) + (\sigma_{i} + \varepsilon)^{-1} \right]$ where $\varepsilon = 10^{-5}$ is a stability constant and $\sigma_{1} \le \ldots \le \sigma_{B}$ are the eigenvalues of the correlation matrix corresponding to the Jacobian $J = \left[\frac{\partial f}{\partial s_{1}},\ldots, \frac{\partial f}{\partial s_{B}} \right]^{\top}$ with $B$ input images $\{s_{1},\ldots, s_{B}\}$.

This metric/predictor has been found to be a strong signal for accuracy in SL NAS among many previous predictors \citep{how_powerful_performance_predictor_nas}. However, for the RL case, just like the loss, the mentioned metrics must be defined with respect to the current replay buffer $\mathcal{D}$, and thus raises the question of what type of data is to be used for calculating these metrics. When using a reasonable variant where the data is collected from a pretrained policy, we found that methods such as Jacobian Covariance did not provide meaningful feedback.


\item Supernet Training: As seen from the results in Figure \ref{fig:degenerated-supernet} (main body) and Appendix \ref{subsec:correlation}, there is a correlation between supernet and discrete cell performances. However, this is affected by the environment used as well as integrality gaps between the continuous relaxation and discrete counterparts, and thus further exploration is needed before concluding that improving the supernet training leads to better discrete cell performances. In any case, reasonable methods of improving the supernet can involve DARTS-agnostic modifications to the RL pipeline, including data augmentation \citep{image_augmentation_all_you_need, autodrac} as well as online hyperparameter tuning \citep{pb2, online_hp_autorl}. Simple hyperparameter tuning (e.g. on the softmax temperature for calculating $p_{o}^{(i,j)}$'s) also can be effective.

\end{enumerate}

\clearpage

\section{Hyperparameters}
\label{appendix:hyperparameters}
In our code, we entirely use Tensorflow 2 for auto-differentation, as well as the April 2020 version of Procgen. For compute, we either used P100 or V100 GPUs based on convenience and availability. Below are the hyperparameter settings for specific methods. For all training curves, we use the common standard for reporting in RL (i.e. plotting mean and standard deviation across 3 seeds).

\subsection{DARTS}
Initially, we sweeped the softmax temperature in order to find a stable default value that could be used for all environments. For PPO, the sweep was across the set $\{1/5.0, 1/10.0, 1/15.0\}$. For Rainbow, the sweep was across $\{1/10.0, 1/20.0, 1/50.0\}$. 

For tabular reported scores in Figures \ref{ppo_clasic_eval} and \ref{rainbow_micro_eval}, we used a consistent softmax temperature of 1/5.0 for PPO, and 1/10.0 for Rainbow.

\subsection{Rainbow-DQN}
We use Acme \citep{acme} for our code infrastructure. We use a learning rate $5 \times 10^{-5}$, batch size 256, n-step size 7, discount factor 0.99. For the priority replay buffer \citep{priority_replay}, we use priority exponent 0.8, importance sampling exponent 0.2, replay buffer capacity 500K. For particular environments (Bigfish, Bossfight, Chaser, Dodgeball, Miner, Plunder, Starpilot), we use n-step size 2 and replay buffer capacity 10K. For C51 \citep{c51}, we use 51 atoms, with $v_{min}=0, v_{max} = 1.0$. As a preprocessing step, we normalize the environment rewards by dividing the raw rewards by the max possible rewards reported in \citep{procgen}.

\subsection{PPO}
We use TF-Agents \citep{TFAgents} for our code infrastructure, along with equivalent PPO hyperparameters found from \citep{procgen}. Due to necessary changes in minibatch size when applying DARTS modules or networks with higher GPU memory usage, we thus swept learning rate across $\{1 \times 10^{-4}, 2.5 \times 10^{-4}, 5 \times 10^{-4}\}$ and number of epochs across $\{1,2,3\}$. 

For all models, we use a maximum power of 2 minibatch size before encountering GPU out-of-memory issues on a standard 16 GB GPU. Thus, for a $16 \times 3 = [16,16,16]$ DARTS supernet, we set the minibatch size to be $256$, which is also used for evaluation with a $64 \times 3 = [64,64,64]$ discretized CNN. Our hyperparameter gridsearch for the evaluation led to an optimal setting of learning rate $=1 \times 10^{-4}$ and number of epochs $=1$.

\subsection{SAC}
We use open-source code found in \url{https://github.com/google-research/pisac}, although we disabled the predictive information loss to use only regular SAC. The baseline architecture is a 4-layer convolutional architecture found in \url{https://github.com/google-research/pisac/blob/master/pisac/encoders.py}. Image-based observations are resized to $64 \times 64$ with a frame-stacking of 3. Both our DARTS supernet and discrete cells use $N=3, I=4, K=1$ using the "Micro" search space, with convolutional depths of $32$ to remain fair to the baseline.

\clearpage
\subsection{Training Procedure}
\label{subsec:training_procedure}
Below are PPO \citep{ppo} and Rainbow-DQN \citep{rainbow_dqn} RL-DARTS variants, which provide an example of the specific training procedure we use. Exact loss definitions and data collection procedures can be found in their respective papers. 

\scalebox{0.85}{
\begin{algorithm}[H]
\SetAlgoLined
Supernet training: \\
\Indp 
Setup supernet encoder $f_{\theta_{e}, \alpha}$ with weights $\theta_{e}$.\\
Initialize policy and value head projection weights $W_{\pi} \in \mathbb{R}^{d, |\mathcal{A}|}, W_{v} \in \mathbb{R}^{d, 1}$. \\
Collect all trainable weights $\theta = \{\theta_{e}, W_{\pi}, W_{v}\}$. \\
Setup policy $\pi_{\theta, \alpha}(s) \sim  \text{softmax}(W_{\pi} \cdot f_{\theta_{e}, \alpha}(s))$. \\
Setup value function $V_{\theta, \alpha}(s) = W_{v} \cdot f_{\theta_{e}, \alpha}(s)$. \\
Define standard PPO loss $\mathcal{L}(\theta, \alpha)$ using $\pi_{\theta, \alpha}$ and $V_{\theta, \alpha}$. \\
Perform PPO training via collecting data from $\pi_{\theta, \alpha}$ and SGD with $\nabla_{\theta, \alpha} \mathcal{L}(\theta, \alpha)$. \\
Collect $\alpha^{*}$ from previous training procedure. \\
\Indm
Discretization: \\
\Indp
Let $\delta(\alpha^{*})$ be the discrete cell constructed via Algorithm \ref{algo:discretization}. \\
\Indm
Evaluation: \\
\Indp
Setup discretized cell encoder $f_{\phi_{e}, \delta(\alpha^{*})}$. \\
Initialize policy and value head projection weights $W_{\pi}' \in \mathbb{R}^{d, |\mathcal{A}|}, W_{v}' \in \mathbb{R}^{d, 1}$. \\
Collect all trainable weights $\phi = \{\phi_{e}, W_{\pi}', W_{v}'\}$. \\
Setup policy $\pi_{\phi, \delta(\alpha^{*})}(s) \sim  \text{softmax}(W_{\pi}' \cdot f_{\phi_{e}, \delta(\alpha^{*})}(s))$. \\
Setup value function $V_{\phi, \delta(\alpha^{*})}(s) = W_{v}' \cdot f_{\phi_{e}, \delta(\alpha^{*})}(s)$. \\
Define standard PPO loss $\mathcal{L}_{\delta(\alpha^{*})}(\phi)$ using $\pi_{\phi, \delta(\alpha^{*})}$ and $V_{\phi, \delta(\alpha^{*})}$. \\
Perform PPO training via collecting data from $\pi_{\phi, \delta(\alpha^{*})}$ and SGD with $\nabla_{\phi} \mathcal{L}_{\delta(\alpha^{*})}(\phi)$. \\
Report final policy reward. \\
\caption{RL-DARTS with PPO}
\label{algo:ppo}
\end{algorithm}
}

\scalebox{0.85}{
\begin{algorithm}[H]
\SetAlgoLined
Supernet training: \\
\Indp 
Setup supernet encoder $f_{\theta_{e}, \alpha}$ with weights $\theta_{e}$.\\
Initialize dueling network projections $W_{v} \in \mathbb{R}^{d, 1}, W_{a} \in \mathbb{R}^{d, |\mathcal{A}|}$. \\
Collect all trainable weights $\theta = \{\theta_{e}, W_{v}, W_{a}\}$. \\
Setup value network $V_{\theta, \alpha}(s) = W_{v} \cdot f_{\theta_{e}, \alpha}(s)$. \\
Setup advantage network $A_{\theta, \alpha}(s, a) = W_{a} \cdot f_{\theta_{e}, \alpha}(s)$. \\
Setup Q-network $Q_{\theta, \alpha}(s, a) = V_{\theta, \alpha}(s) + A_{\theta, \alpha}(s,a) - \frac{1}{|\mathcal{A}|} \sum_{a' \in \mathcal{A}} A_{\theta, \alpha}(s, a')$. \\
Define standard Rainbow loss $\mathcal{L}(\theta, \alpha)$ using $Q_{\theta, \alpha}$. \\
Perform Rainbow training via collecting data from $Q_{\theta, \alpha}$ and SGD with $\nabla_{\theta, \alpha}\mathcal{L}(\theta, \alpha)$. \\
Collect $\alpha^{*}$ from previous training procedure.

\Indm
Discretization \\ 
\Indp
Let $\delta(\alpha^{*})$ be the discrete cell constructed via Algorithm \ref{algo:discretization}. \\
\Indm
Evaluation \\
\Indp

Setup discretized cell encoder $f_{\phi_{e}, \delta(\alpha^{*})}$ with weights $\phi_{e}$.\\
Initialize dueling network projections $W_{v}' \in \mathbb{R}^{d, 1}, W_{a}' \in \mathbb{R}^{d, |\mathcal{A}|}$ \\
Collect all trainable weights $\phi = \{\phi_{e}, W_{v}', W_{a}'\}$ \\
Setup value network $V_{\phi, \delta(\alpha^{*})}(s) = W_{v}' \cdot f_{\phi_{e}, \delta(\alpha^{*}) }(s)$ \\
Setup advantage network $A_{\phi, \delta(\alpha^{*})}(s, a) = W_{a}' \cdot f_{\phi_{e}, \delta(\alpha^{*})}(s) $ \\
Setup Q-network $Q_{\phi, \delta(\alpha^{*})}(s, a) = V_{\phi, \delta(\alpha^{*})}(s) + A_{\phi, \delta(\alpha^{*})}(s,a) - \frac{1}{|\mathcal{A}|} \sum_{a' \in \mathcal{A}} A_{\phi, \delta(\alpha^{*})}(s, a')  $ \\
Define standard Rainbow loss $\mathcal{L}_{\delta(\alpha^{*})}(\phi)$ using $Q_{\phi, \delta(\alpha^{*})}$. \\
Perform Rainbow training via collecting data from $Q_{\phi, \delta(\alpha^{*})}$ and SGD with $\nabla_{\phi}\mathcal{L}_{ \delta(\alpha^{*})}(\phi)$. \\
Report final policy reward. \\
\Indm
\caption{RL-DARTS with Rainbow. Note that we do not use noisy nets in this implementation.}
\label{algo:rainbow}
\end{algorithm}
}

\clearpage

\begin{algorithm}[h]
\SetAlgoLined
Argmax: \\
\Indp
For $(i,j)$ across all edges: \\
\Indp
Define edge strength $w_{i,j} = \max_{o \in \mathcal{O}, o \neq zero} p_{o}^{(i,j)}$. \\
Define edge op $o_{(i,j)} = \argmax_{o \in \mathcal{O}, o \neq zero} p_{o}^{(i,j)}$. \\
\Indm
\Indm

Prune: \\
\Indp
For node $j$ in all intermediate nodes: \\
\Indp
Sort input edge weights $w_{i_{1}, j} \geq w_{i_{2}, j} \geq \ldots $ \\ 
Retain only top $K$ edges $(i_{1}, j), \ldots,(i_{K}, j)$ and corresponding ops $o_{(i_{1}, j)}, \ldots,o_{(i_{K}, j)}$ in final cell.\\

\caption{Discretization Procedure.}
\label{algo:discretization}
\end{algorithm}

Note that both RL-DARTS procedures can also be summarized in terms of raw code as simple one-line edits to the image encoder used (compressing the rest of the regular RL training pipeline code):

\begin{verbatim}
def train(feature_encoder):
    """Initial RL algorithm setup"""
    ...
    extra_variables = Wrap(feature_encoder)
    all_trainable_variables = 
       [feature_encoder.trainable_variables(), extra_variables]
    """Rest of RL algorithm setup"""
    ...
    apply_gradients(loss, all_trainable_variables)
\end{verbatim}

Thus, the 3-step RL-DARTS procedure from Section \ref{sec:rl_darts_method} can be seen as:
\begin{verbatim}
DARTSSuperNet = MakeSuperNet(ops, num_nodes) # Setup
train(DARTSSuperNet) # Supernet training
DiscretizedNet = DARTSSuperNet.discretize() # Discretization
train(DiscretizedNet) # Evaluation
\end{verbatim}

\clearpage

\section{Miscellaneous}
\subsection{Search Space Size}
\label{appendix:search_space_size}
Let $O_{nz} = |\mathcal{O}| - 1$, the number of non-zero ops in $\mathcal{O}$. For a cell (normal or reduction), the first intermediate node can only be connected to the input via a single op, and thus the choice is only $O_{nz}$. However, later intermediate nodes use $K=2$ inputs which leads to a choice size of $O_{nz}^{K} \times {i \choose K}$ where $i$ is the index of the intermediate node. Thus the total number of possible discrete cells is $O_{nz} \cdot \prod_{i=2}^{I} \left(O_{nz}^{K} \times {i \choose K} \right)$.

For the "Classic" search space, there are both normal and reduction cells to be optimized, with number of non-zero normal ops $O_{nz, N} = 5$ and number of non-zero reduction ops $O_{nz, R} = 4$, with $I=4, K=2$ for both. This leads to a total configuration size of  $\left[O_{nz, N} \cdot \prod_{i=2}^{4} \left(O_{nz,N}^{2} \times {i \choose 2} \right) \right] \times \left[O_{nz, R} \cdot \prod_{i=2}^{4} \left(O_{nz, R}^{2} \times {i \choose 2} \right) \right] \approx 4 \times 10^{11}$.

For the "Micro" search space, since we do not use reduction cells in order to simplify visualizations and ablation studies, $O_{nz, N}=4$ with $I=4, K=2$. This gives $\left[O_{nz, N} \cdot \prod_{i=2}^{4} \left(O_{nz}^{2} \times {i \choose 2} \right) \right] \approx 3 \times 10^{5}$, which is comparable to the search space of size $5^{6} \approx 1.5 \times 10^{4}$ in NASBENCH-201 \citep{nasbench_201}.